\documentclass{article} 
\usepackage{iclr2027_conference,times}
\iclrfinalcopy

\usepackage{amsmath,amsfonts,bm}

\def\eqref#1{equation~\ref{#1}}

\def\1{\bm{1}}

\DeclareMathAlphabet{\mathsfit}{\encodingdefault}{\sfdefault}{m}{sl}
\SetMathAlphabet{\mathsfit}{bold}{\encodingdefault}{\sfdefault}{bx}{n}

\usepackage{hyperref}
\usepackage{url}
\usepackage{graphicx}
\usepackage{amssymb}

\title{Emergent phases of superposition: from partial to full representation}

\author{Lihao Guo, Yizhou Liu \& Jeff Gore \\
Massachusetts Institute of Technology\\
Cambridge, MA 02139, USA \\
\texttt{\{glh123,liuyz,gore\}@mit.edu}
}

\begin{document}

\maketitle

\begin{abstract}
  Large language models are thought to represent features by vectors in a hidden space of dimension given by the model's width. Superposition, in which more features are represented than the width by letting representation vectors overlap, is a leading account of how representation vectors are organized. However, how model width and data statistics determine the configuration of representation vectors and the resulting loss when the number of features and the width are large remains less understood. Here we show, in Anthropic's toy model of superposition, that increasing the width drives a continuous phase transition from a partial-representation phase, where only a subset of features receives appreciable representation vectors while the rest vanish, to a full-representation phase, where every feature is represented. Our theory via a partial random projection approximation predicts, and experiments confirm, that the critical width grows linearly with the number of active features up to a logarithmic factor. The loss scaling changes across the transition: below the critical width, the loss grows linearly with the number of active features and depends weakly on the width in a form set by data statistics; above it, the loss grows approximately quadratically with the number of active features and decays inversely with the width. Non-uniform firing probabilities delay the transition and lower the loss, as more frequent features occupy more space. Our results provide an account of how model width and data statistics jointly shape representations and loss, a step toward understanding representation scaling in large models.
\end{abstract}

\section{Introduction}

Large language models (LLMs) have achieved remarkable empirical success across a wide range of tasks, yet the internal mechanism behind this success remains unclear. A leading view is that LLMs compute with features. Features are interpretable properties of the input, and serve as the fundamental units of concept representation and computation \citep{elhage_toy_2022,bricken_towards_2023}. Under the linear representation hypothesis, each feature corresponds to a direction in hidden space (its representation vector), and the hidden state of an input is approximately a linear combination of the representation vectors of its active features \citep{mikolov_linguistic_2013,park_linear_2023}. Consistent with this picture, dictionary-learning methods have extracted millions of such feature directions from LLM hidden layers \citep{bricken_towards_2023,templeton_scaling_2024}. How these representation vectors are organized in hidden space is thus a central question for understanding LLMs.

How representation vectors are organized, and how that organization sets performance, has been widely studied. A linear network with hidden dimension $m$ (the model width) can represent at most $m$ features orthogonally, and its optimum is a principal-component projection onto the most important ones; the rest are discarded \citep{baldi_neural_1989,elhage_toy_2022}. Nonlinearity lifts this ceiling: the network can pack more features than dimensions by letting representation vectors overlap and using the nonlinear layer to filter the resulting interference, a mechanism called superposition \citep{elhage_toy_2022}. Superposition has since been characterized in two regimes. When the numbers of active features and hidden dimensions are both of order one, the representation configuration is determined by the geometry of a few vectors, which settle into specific configurations such as antipodal pairs and polytopes \citep{elhage_toy_2022,chen_dynamical_2023}. In the opposite, thermodynamic limit, existing analyses take the picture to be statistical: every feature is assigned a representation vector, the vectors are treated as nearly orthogonal, and their overlaps set a loss that decays inversely with the width \citep{cowsik_persian_2024,liu_superposition_2025}. What has not been established is whether this picture holds across model widths and data statistics: whether a model in this limit can instead leave some features unrepresented, when it does so, and how the loss changes across that boundary. This leads to the question we address: \textbf{How do model width and data statistics determine the representation behaviors (which features are represented, with what representation vector) and the resulting loss?}

We address this question in the toy model of superposition introduced by Anthropic \citep{elhage_toy_2022}, varying model width and data statistics. In trained models, we identify two types of configurations of the representation vectors: a partial representation at small model width $m$, where features split into a group with appreciable norms and a group with vanishing norms (even though all features are uniform), and a full representation at large width, where every feature carries an appreciable vector. As $m$ grows, the number of represented features increases continuously until full representation is reached, and the model undergoes a phase transition from partial representation to full representation. We develop an approximation of the representation behavior and the loss, which reproduces the trends of trained models and captures their asymptotic behavior in the thermodynamic limit. The loss follows the representation, showing different behaviors on two sides of the transition. Finally, we characterize how sparsity and non-uniformity shift the transition and the loss.

\begin{figure}[tbp]
\begin{center}
\includegraphics[width=\textwidth]{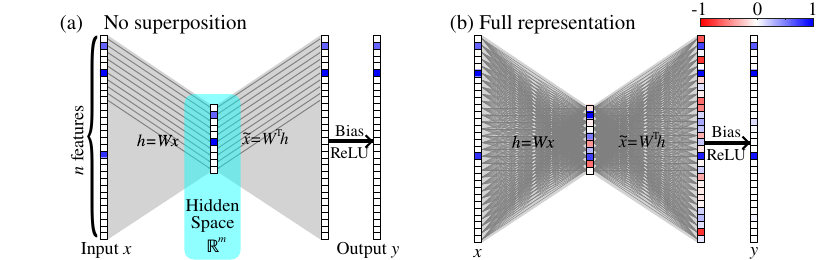}
\end{center}
\caption{The toy model composed of embedding and unembedding layers exhibits different types of representation configurations. (a) and (b) are schematic illustrations of the toy model. Colored boxes are used to represent the element values of the input, hidden, and output vectors. (a) No superposition. Only $m$ features are represented by vectors without overlap, and the remaining features are discarded. (b) Full representation. All features are represented by vectors with appreciable norms.}
\label{fig:intro-model}
\end{figure}

\section{Toy model}
\label{sec:toy-model-its}

We adopt the toy model of superposition \citep{elhage_toy_2022}, see Fig. \ref{fig:intro-model}. In this model, an input $\boldsymbol x\in \mathbb R^n$ is mapped to a lower-dimensional vector $\boldsymbol h$ in the hidden space $\mathbb R^m$ through matrix $\boldsymbol  W\in \mathbb  R^{{m\times n}}$, and then unembedded:
\begin{align}
\label{eq:Model-arch}
  \boldsymbol h&\equiv \boldsymbol W \boldsymbol x, ~~~~~~~~~~  {\boldsymbol y}\equiv \left [\boldsymbol W^T\boldsymbol h + \boldsymbol b \right ]_+,
\end{align}
where $[z]_+\equiv \max(z,0)$ is the ReLU operator, and $\boldsymbol b\in \mathbb R^n$ is the bias vector. The hidden space has lower dimensionality than the input space, acting as the bottleneck, hence $m$ is also called the model width. The loss is defined as the mean square error between the output $\boldsymbol y$ and the input $\boldsymbol x$:
\begin{equation}
  \label{eq:Model-loss}
  \mathcal L \equiv  {1\over n} \mathbb E_{\boldsymbol x} \| \boldsymbol y-\boldsymbol x\|^2.
\end{equation}
Unlike \citet{elhage_toy_2022}, we assign equal importance to all features. During training, the model aims to minimize $\mathcal L$ by optimizing $\boldsymbol W$ and $\boldsymbol b$. See appendix \ref{sec:toy-model-training} for training details.

The input $\boldsymbol x$ is interpreted as a list of activations, whose $i$th element $x_i$ is the activation of feature $i$. Under this interpretation, $n$ is the total number of features. The activations follow
\begin{equation}
\label{eq:Model-input}
x_i = u_i v_i,
\end{equation}
where $u_i$ and $v_i$ are all independent. The random variable $u_i$ follows Bernoulli distribution $\mathcal B(p_i)$, where $p_i$ is the firing probability of feature $i$. It controls whether a feature is active (non-zero) or not. Random variables $v_i$ control activation strengths of active features. They are i.i.d. random variables following a distribution $\mathcal D$ with mean $\bar v$ and variance $\sigma _v^2$. In this paper, for simplicity, we let $\mathcal D$ be the uniform distribution $\mathcal U[\bar v-\sqrt{ 3 }\sigma _v,\bar v+\sqrt{ 3 }\sigma _v]$ and let $\bar v=1,\sigma _{v}={1\over \sqrt{ 3}}$ (i.e., $v\sim \mathcal U[0,2]$ as in \citet{liu_superposition_2025}). The effects of $\bar v$ and $\sigma _v$ are discussed in appendix \ref{sec:effect-sigma-_v}. The input is sparse in the sense that most elements are zero. Denoting the number of active features by $E\equiv \sum_iu_i$, we use its mean $\bar E$ to quantify the sparsity of the input ensemble; it satisfies $\bar E\ll n$.

Under the feature interpretation, the embedding matrix $\boldsymbol W$ can be considered as $n$ vectors:
\begin{equation}
\label{eq:def-feature-vecs}
\boldsymbol W=(\boldsymbol w_1~\boldsymbol w_2~...~\boldsymbol w_n),
\end{equation}
where $\boldsymbol w_i$ is the representation vector of feature $i$ in the hidden space $\mathbb R^m$. Their norms are denoted by $s_i\equiv \| \boldsymbol w_i\|$. Depending on the data properties and the model width $m$, some features may not be represented at the loss minimum, i.e., the norms of their representation vectors are close to or exactly 0. We define the effective number of represented features as the participation ratio of norms:
\begin{equation}
\label{eq:def-simulation-r}
r\equiv {\left (\sum_i s_i ^2\right )^2 \over \sum_i s_i ^4}.
\end{equation}
The participation ratio $r$ ranges from 1 to $n$, counting vectors with appreciable norm while ignoring near-zero ones; if $k$ vectors share the same norm and the remaining vectors vanish, $r=k$. When $r\le m$, vectors of represented features can be orthogonal and be free of superposition. The corresponding loss is given in appendix \ref{sec:non-superp-loss}. When $m<r<n$, vectors have to overlap and superposition emerges, but some features are sacrificed and the representation is partial. When $r=n$, all features are represented (full representation).

\section{Results}
\label{sec:results}

In Sec. \ref{sec:trans-from-part-2-full}, we discuss the effect of model width $m$ on the representation behavior and the loss, and show the phase transition from partial to full representation as $m$ increases. Based on the phenomena, we develop the partial random projection (RP) approximation to estimate the representation behavior and the loss. In Sec. \ref{sec:spars-isotr-indep}, we apply the approximation to study the effects of sparsity. In Sec. \ref{sec:non-unif-indep}, we further introduce non-uniformity to firing probabilities, and study its effects.

\subsection{Width drives a transition from partial to full representation}
\label{sec:trans-from-part-2-full}

\begin{figure}[tb]
\begin{center}
\includegraphics[width=0.95\textwidth]{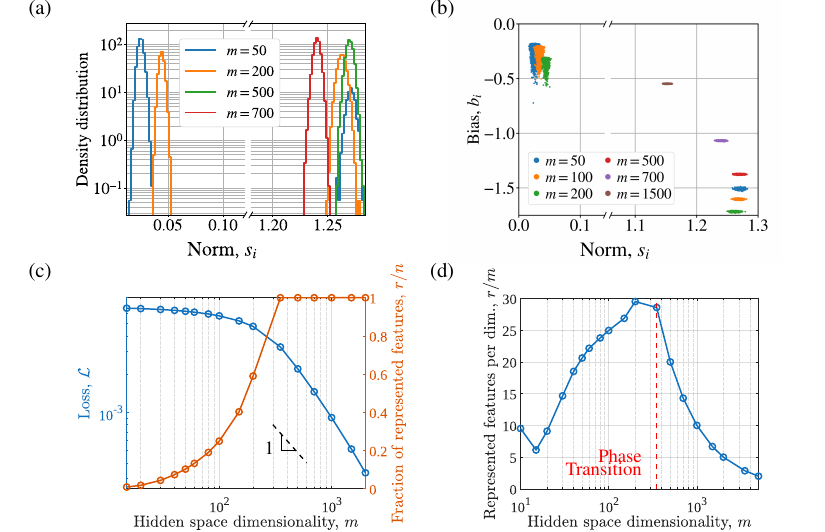}
\end{center}
\caption{As the width grows, the model undergoes a transition from partial to full representation. (a) Distributions of $s_i$ at different $m$. At small $m$ the distribution is bimodal while at large $m$ the distribution is unimodal. (b) The distribution of $s_i$ and $b_i$ of each feature. Sample points at small $m$ form two clusters while at large $m$ they form one cluster. (c) The loss $\mathcal L$ and the fraction $r/n$ of represented features as functions of $m$. (d) The ratio $r/m$ as a function of $m$. It is always greater than 1, indicating superposition. $n=10000,\bar E=50$.}
\label{fig:transition}
\end{figure}

In this section, we focus on how model width $m$ influences the representation behavior and the loss $\mathcal L$. For this, we simply let all features have the same firing probability $p\equiv {\bar E\over n}$.

Given $n$ and $\bar E$, the converged representation of the trained model (for $3\times 10^4$ steps; appendix \ref{sec:toy-model-training}) shows different behaviors at different $m$. At small $m$, features are split into two groups. In one group, features have nearly vanishing $s_i$, and are essentially unrepresented. In another group, features have $s_i$ of order 1 and are well represented. Norms $s_i$ are nearly equal within each group. At large $m$, all features have appreciable representation vectors with nearly equal $s_i$ (Fig. \ref{fig:transition}a). Biases behave in the same way as norms: At small $m$, biases are nearly equal within each group; at large $m$, all biases are nearly equal (Fig. \ref{fig:transition}b). As $m$ increases, more features are represented, until full representation is reached and a phase transition happens. We define the critical width $m^{*}$ as the minimal $m$ at which the distribution of $s_i$ becomes unimodal.

Correspondingly, the loss $\mathcal L$ also shows different behaviors on two sides of the transition. Before the transition, the loss $\mathcal L$ decays slowly with $m$; after the transition, the loss decays approximately as $1/m$ (Fig. \ref{fig:transition}c). The $\mathcal L$-$m$ curve bends around $m^{*}$.

Note that the partial representation before the transition does not mean no superposition. The number $r$ of features represented is always greater than $m$ (Fig. \ref{fig:transition}d), and $r/m$ peaks around the transition. In what follows we compare $r$ with $n$ (completeness of representation), not with $m$. In summary:
\begin{center}
\fbox{\parbox{0.92\linewidth}{%
\textbf{Key result 1:} As the width $m$ increases, the model undergoes a phase transition from partial representation to full representation.
}}
\end{center}

We use the term phase transition in the following sense. The fraction $r/n$ serves as the order parameter: It is below 1 for $m< m^{*}$ and equal to 1 for $m>m^{*}$, and it reaches 1 continuously, so the transition is continuous. The loss is smooth on either side of $m^{*}$, but its dependence on m changes at $m^{*}$. Strictly, a sharp transition requires the limit $1\ll \bar E, m \ll n$, in which $m$ can be treated as continuous.

\begin{figure}[tb]
\begin{center}
\includegraphics[width=0.97\textwidth]{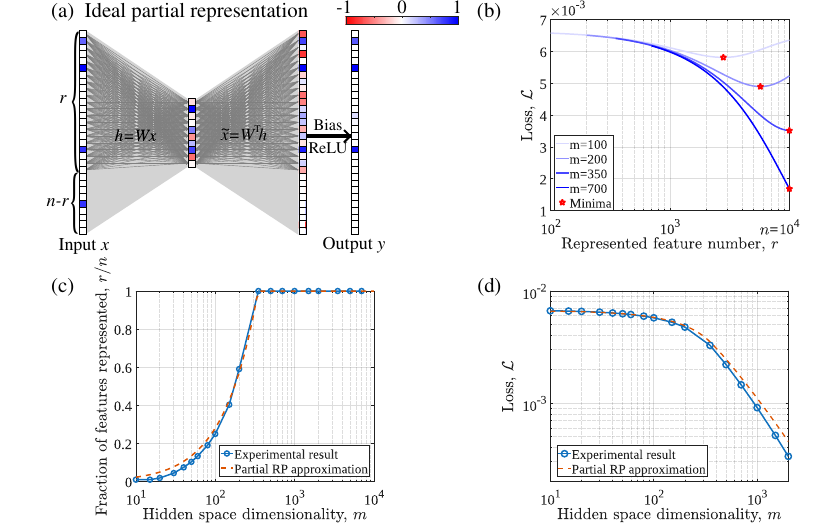}
\end{center}
\caption{The partial random projection approximation captures the behavior of toy model. (a) Schematic illustration of ideal partial representation. (b) The loss $\mathcal L$ predicted by partial RP approximation as a function of $r$ at different $m$. (c) The predicted $r_{{\rm opt } }$ shows good consistency with the experimental $r$. (d) The predicted $\mathcal L$ shows good consistency with experimental loss. $n=10000, \bar E=50$.}
\label{fig:analytic-solution}
\end{figure}

To understand these phenomena, we develop an approximation to $\mathcal L$, which we call the partial random projection (RP) approximation. We idealize the representation vectors $\boldsymbol w_i$ as forming two groups. One group contains vectors of $r$ represented features, each with the same norm $s_i=s$. The other group contains zero vectors of $(n-r)$ unrepresented features (Fig. \ref{fig:analytic-solution}a). In the represented group, biases $b_i=b$ are also assumed equal. We apply this approximation to both partial and full representation phases, as we expect the unrepresented group to vanish spontaneously at the loss minimum in the regime of full representation. The loss is thus composed of the contributions of two groups:
\begin{equation}
  \label{eq:loss-total}
\mathcal L(\bar E,m,n;r)={1\over n}\left ( r \mathcal L_{{\rm rep } } + (n-r)\mathcal L_{{\rm unrep } } \right ).
\end{equation}
The loss $\mathcal L_{{\rm unrep } }$ contributed by unrepresented features is given in appendix \ref{sec:estim-part-repr-w-part-RP-approx}:
\begin{equation}
  \label{eq:loss-unrep}
\mathcal L_{{\rm unrep } }=(\bar v^2+\sigma _v^2){\bar E\over n}-\bar v^2{\bar E^2\over n^2}.
\end{equation}
Given an input $\boldsymbol x$, the loss $\mathcal L_{{\rm rep } }$ contributed by represented features can be further divided into two terms:
\begin{equation}
\mathcal L_{{\rm rep } }|_{\boldsymbol x}={1\over r}\left \{ \sum_{i=1}^{{E_r}} \left [ \left [ s^2v_{a_i} + s^2 \eta _{a_i}+b \right ]_+ -v_{a_i} \right ]^2+\sum_{i\not\in \{a_k \}}[s^2\eta _i +b ]_+^2 \right \},
\end{equation}
where the first term is the loss contribution of active features, and the second term is that of inactive features. The set $\{a_k\}$ contains all active features within the represented group, and $E_r$ is the number of them. The crosstalk noise on feature $i$ is
\begin{equation}
\eta _i \equiv \sum_{j\in \{a_k\} \backslash i} {\boldsymbol w_i\cdot \boldsymbol w_j\over s^2}  v_j.
\end{equation}
We treat $\eta _i$ as random variables independent of $i$ and of $\boldsymbol v$ by mean field approximation, and further use a Gaussian distribution to approximate it:
\begin{equation}
\label{eq:xtalk-noise-approx-distribution}
\eta _i \sim \mathcal N \left (0, E_r(\bar v^2+\sigma _v^2){r-m \over m(r-1)}\right ).
\end{equation}
The variance of $\eta _i$ is factorized into three pieces: the number of active represented features $E_r$, the activation second moment $(\bar v^2+\sigma _v ^2)$, and the minimal mean squared overlap ${r-m \over  m(r-1)}$ confined by the Welch bound \citep{welch_lower_1974}. For $m\ll r$ the overlap approaches $1\over m$, the value for random projection vectors, which is the origin of the name. Following Eq. (\ref{eq:xtalk-noise-approx-distribution}), the loss $\mathcal L_{{\rm rep } }$ averaged over $\boldsymbol x$ ensemble can be approximated by
\begin{align}
  \mathcal L_{{\rm rep } }&= \min_{s,b} \mathbb E_{\boldsymbol x} \mathcal L_{{\rm rep } }|_{\boldsymbol x}   \approx  \min_{s,b} {1\over r} \mathbb E_{E_r }\left \{ E_r \mathbb E_{\eta,v } \left [[s^2v+s^2\eta +b]_+-v\right ]^2 +(r-E_r)\mathbb E_{\eta } [s^2\eta +b]_+^2 \right \}, \label{eq:loss-rep}
\end{align}
where $E_r$ follows a binomial distribution: $ p(E_r) = \binom{r}{E_r} p^{E_r} (1-p)^{r-E_r}$. Plugging Eq. (\ref{eq:loss-unrep}) and (\ref{eq:loss-rep}) into Eq. (\ref{eq:loss-total}), we obtain the loss $\mathcal L(\bar E,m,n;r)$ with $r$ still tunable. The predicted loss is obtained by further optimization with respect to $r\in [m,n]$:
\begin{align}
  \mathcal L(\bar E,m,n)&=\min_r \mathcal L(\bar E,m,n;r), ~~~~  r_{{\rm opt } }(\bar E,m,n)\equiv \arg\min_r \mathcal L(\bar E,m,n;r). \label{eq:def-r-opt}
\end{align}

The partial RP approximation captures the transition from partial to full representation. The optimal $r_{{\rm opt } }$ grows with $m$, while it is always greater than $m$ (vectors are in superposition, see appendix \ref{sec:subopt-non-superp}). The transition arises because $r$ moves from the interior of $(m,n)$ to $n$  (Fig. \ref{fig:analytic-solution}bc). The loss trend and the bending around $m^{*}$ are also predicted (Fig. \ref{fig:analytic-solution}d). At finite $\bar E$ and $n$, the predicted loss in the full-representation phase is slightly higher than the experimental loss due to the finite size effect, which disappears in the limit $1\ll \bar E \ll n$ (appendix \ref{sec:finite-size-effect}).

\subsection{Sparsity sets the critical width and the loss}
\label{sec:spars-isotr-indep}

\begin{figure}[tbp]
\begin{center}
\includegraphics[width=0.97\textwidth]{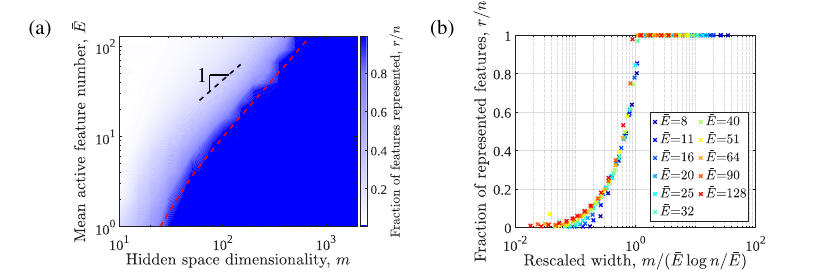}
\end{center}
\caption{The critical width $m^{*}$ increases with the active feature number $\bar E$, growing asymptotically proportional to it up to a logarithmic correction. (a) Phase diagram of $r/n$ as a function of $\bar E$ and $m$. The red dashed curve is the phase boundary predicted by partial RP approximation. (b) By rescaling $m$ according to Eq. (\ref{eq:asymptotic-trans-m}), the critical points collapse. $n=10000$.}
\label{fig:trans-m-vs-E}
\end{figure}

\begin{figure}[tb]
\begin{center}
\includegraphics[width=0.97\textwidth]{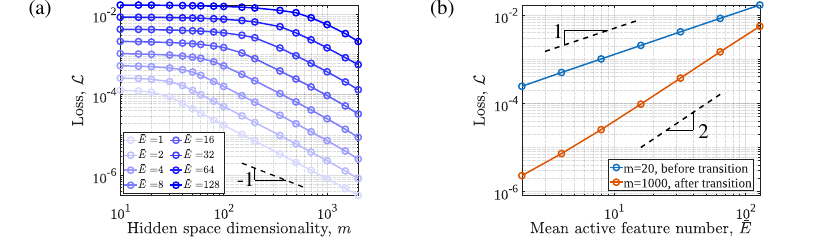}
\end{center}
\caption{The loss grows with number of active features, approximately proportionally in partial representations and quadratically up to a logarithmic correction in full representations. (a) As $\bar E$ increases, the $\mathcal L$-$m$ curve shifts upwards. (b) The loss $\mathcal L$ as a function of $\bar E$ under fixed $n$ and two different $m$ (before and after transition). $n=10000$.}
\label{fig:loss-vs-E}
\end{figure}

We now study how sparsity determines the representation behavior and the loss, using both the converged trained model and the partial RP approximation. The sparsity is measured by the number of active features $E$ in each input (smaller $E$ means sparser input). Here we only focus on its mean value $\bar E$, and the discussion of its distribution can be found in appendix \ref{sec:deta-deriv-loss-exch-corr-feat}. To isolate the effect of $\bar E$, we keep the uniform setup in Sec. \ref{sec:trans-from-part-2-full}. The effect of sparsity is summarized as follows (Fig. \ref{fig:trans-m-vs-E} and \ref{fig:loss-vs-E}):
\begin{center}
\fbox{\parbox{0.92\linewidth}{%
\textbf{Key result 2:} The critical width $m^{*}$ increases with the mean active feature number $\bar E$. In the limit $1\ll \bar E \ll n$, it grows asymptotically as $m^{*}\sim \bar E \log {n\over \bar E}$. The loss $\mathcal L$ grows with $\bar E$. When $m\ll m^{*}$, the loss $\mathcal L \approx {\bar E\over n}(\bar v^2+\sigma _v^2)$; when $m\gg m^{*}$, the loss grows asymptotically as $\mathcal L\approx  {2\sigma _v^2\bar E^2\over m n} \log {n\over \bar E}$.
}}
\end{center}

We use the partial RP approximation to understand the result. Increasing $\bar E$ raises $\mathcal L$ through both represented and unrepresented features. The loss of unrepresented features (\ref{eq:loss-unrep}) is proportional to $\bar E$ in the limit $\bar E\ll n$. For represented features, the distribution of $E_r$ shifts upwards as $\bar E$ increases, hence the crosstalk noise $\eta $ increases, so $\mathcal L_{{\rm rep } }$ increases. We can derive the asymptotic behavior of $\mathcal L_{{\rm rep } }$ in the limit $1\ll  rp \ll m\ll r$ (appendix \ref{sec:large-width-limit}):
\begin{equation}
\label{eq:asymptotic-loss}
\mathcal L_{{\rm rep }}\approx {2\sigma _v^2 p^2 r \over m}\log {1 \over p}={2\sigma _v^2 \bar E^2 r \over n^2 m}\log {n \over \bar E}.
\end{equation}
In full representations (i.e., $r=n$), Eq. (\ref{eq:asymptotic-loss}) gives the total loss:
\begin{equation}
  \label{eq:asymptotic-total-loss}
\mathcal L=\mathcal L_{{\rm rep } }\approx {2\sigma _v^2 \bar E^2 \over n m}\log {n \over \bar E}.
\end{equation}
The critical width $m^{*}$ can now be estimated by comparing the two options of a feature. If it is not represented, it costs $\mathcal L_{{\rm unrep } }$ in Eq. (\ref{eq:loss-unrep}), which is independent of $m$; if it is represented, it costs $\mathcal L_{{\rm rep } }$ in Eq. (\ref{eq:asymptotic-total-loss}), which grows with $\bar E$ faster than $L_{{\rm unrep } }$, but decays with $m$. At small $m$ the ratio $\mathcal L_{{\rm rep } }/\mathcal L_{{\rm unrep } }$ is large and features are unrepresented. The phase transition happens when $\mathcal L_{{\rm rep } }/\mathcal L_{{\rm unrep } }$ becomes of order 1, i.e., at
\begin{equation}
  \label{eq:asymptotic-trans-m}
m^{*}\sim \bar E \log {n \over \bar E},
\end{equation}
with an $O(1)$ prefactor depending on $\sigma _v$ (appendix \ref{sec:large-width-limit}). The prediction of $m^{*}$ is consistent with experimental result (Fig. \ref{fig:trans-m-vs-E}b). This scaling coincides with the classical compressed-sensing bound $m = O(k \log{n\over k})$ for recovering $k$-sparse signals from $m$ random linear measurements \citep{donoho_compressed_2006}, which \citet{elhage_toy_2022} and \citet{klindt_unifying_2026} invoke for superposition. Compressed sensing offers a second reading of Eq. (\ref{eq:asymptotic-trans-m}): the width at which a fixed dictionary ceases to be decodable by a single thresholding step is also the width below which an optimizing model no longer finds full representation worthwhile \citep{ivanitskiy_towards_2026}.

At $m\ll m^{*}$, the fraction $r/n$ is small, hence $\mathcal L$ is dominated by $\mathcal L_{{\rm unrep } }$, growing linearly in $\bar E$; at $m^{*}\ll m \ll n$, the loss $\mathcal L$ follows Eq. (\ref{eq:asymptotic-total-loss}), growing quadratically with $\bar E$ up to a logarithmic correction. These results match the prediction in \citet{cowsik_persian_2024}. The $1/m$ dependence of $\mathcal L$ in full-representation phase also matches the prediction in \citet{liu_superposition_2025}. These asymptotic behaviors all match the experiments well (Fig. \ref{fig:loss-vs-E}b).

\subsection{Non-uniformity delays the transition and lowers the loss}
\label{sec:non-unif-indep}

We now study how the non-uniformity of features' firing probabilities influences the representation behavior and the loss. In this section, the firing probabilities $p_i$ are no longer uniform. Here we use the conventional power-law firing probabilities $p_i\propto i^{-\epsilon _p}$ with the prefactor fixed by $\bar E$, hence features are indexed in decreasing order. The data exponent $\epsilon _p$ controls the degree of non-uniformity, as $p_i$ becomes more uneven when $\epsilon _p$ grows from 0. We only focus on moderate $\epsilon _p$ in this section. Discussions on the applicable range of $\epsilon _p$ can be found in appendix \ref{sec:det-exp-results-indep-feat-with}. 

\begin{figure}[tb]
\begin{center}
\includegraphics[width=0.97\textwidth]{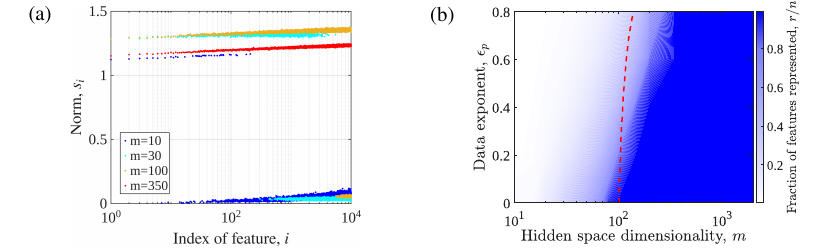}
\end{center}
\caption{The critical width $m^{*}$ increases with non-uniformity, i.e., the data exponent $\epsilon _p$. (a) Distributions of $s_i$ at different $m$. As $m$ increases, two $s_i$ bands change into one band. $\epsilon_p=0.2$ (b) Phase diagram of $r/n$ as a function of $\epsilon _p$ and $m$. The red dashed curve is the phase boundary predicted by partial RP approximation. $n=10000, \bar E=10$.}
\label{fig:non-uniformity}
\end{figure}

\begin{figure}[htbp]
\begin{center}
\includegraphics[width=0.97\textwidth]{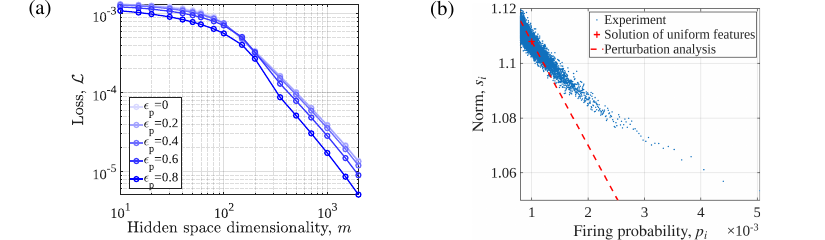}
\end{center}
\caption{The loss decays with the data exponent $\epsilon _p$, since the representation adapts to the non-uniformity. (a) As $\epsilon _p$ increases, the $\mathcal L$-$m$ curve shifts downwards. (b) The norm decreases with $p_i$ in the direction predicted. The red cross point is the solution of uniform features given the same $\bar E,m$ and $n$. $\epsilon _p=0.2, m=1000$. For (a) and (b), $n=10000, \bar E=10$.}
\label{fig:non-uniformity-loss}
\end{figure}

Under the setup of non-uniform independent features, the transition from bimodal $s_i$ distribution (partial representation) to unimodal $s_i$ distribution (full representation) remains valid for moderate $\epsilon _p$. Biases also still behave in the same way as norms (appendix \ref{sec:det-exp-results-indep-feat-with}). More specifically, in partial representations, features with higher $p_i$ are more likely to be represented. Unlike the uniform case, vectors of represented features have non-uniform norms depending on $p_i$ (Fig. \ref{fig:non-uniformity}a). Nevertheless, we still use $r/n$ as the order parameter, since its singular behavior around the critical point persists. We have the following result (Fig. \ref{fig:non-uniformity} and \ref{fig:non-uniformity-loss}):
\begin{center}
\fbox{\parbox{0.92\linewidth}{%
\textbf{Key result 3:} As the data exponent $\epsilon _p$ increases, the critical width $m^{*}$ increases, and the loss $\mathcal L$ decreases.
}}
\end{center}

We adapt the partial RP approximation based on the observations. The represented group is idealized as containing only features with largest $p_i$. When $m$ increases, features are represented in turn from the first one (largest $p_i$) to the last one (smallest $p_i$). Eq. (\ref{eq:loss-unrep}) is then replaced by:
\begin{equation}
  \label{eq:loss-unrep-non-uni}
  \mathcal L_{{\rm unrep } }={1\over n-r} \sum_{i=r+1}^n (\bar v^2+\sigma _v^2)p_i -\bar v^2 p_i^2.
\end{equation}
We still apply the random projection approximation to represented features and formally express $\mathcal L_{{\rm rep } }$ by Eq. (\ref{eq:loss-rep}), but here $E_r$ follows a Poisson binomial distribution generated by $p_1,p_2,...,p_r$. The adapted partial RP approximation captures the trend of $m^{*}$ (Fig. \ref{fig:non-uniformity}b).

The partial RP approximation treats all represented features isotropically. As a result, in the limit $1\ll \bar E\ll m \ll n$, it still gives the asymptotic expression (\ref{eq:asymptotic-total-loss}) which only depends on mean firing probability, predicting no dependence on $\epsilon _p$. This conflicts with the experimental result, as $\mathcal L$ decays with $\epsilon _p$. To explain it, we need to consider the anisotropic organization of the represented features. For each feature, we define the leverage score for represented features \citep{scherlis_polysemanticity_2025}
\begin{equation}
  \label{eq:def-leverage-score}
  h_i\equiv{1\over s_i^2} \boldsymbol w_i^T (\tilde {\boldsymbol W}\tilde {\boldsymbol W}^T)^{-1} \boldsymbol w_i,
\end{equation}
where $\tilde {{\boldsymbol W}}=(\boldsymbol w_1/s_1 ~ ... ~ \boldsymbol w_r/s_r)$ contains normalized vectors of represented features. We set $h_i=0$ for unrepresented features by convention. Here $h_i$ acts as the effective number of hidden dimensions allocated to feature $i$. However, it differs from the feature dimensionality used in \citet{elhage_toy_2022}, as $h_i$ upper-bounds their dimensionality \citep{ivanov_spectral_2026}. In the limit $1\ll m \ll r$, the squared overlaps of feature $i$ with other features are controlled by $h_{i}$ (appendix \ref{sec:pert-analys-indep}):
\begin{equation}
  \sum_{j:j\ne i}{ (\boldsymbol w_i \cdot \boldsymbol w_j)^2 \over s_i^2 s_j^2} \approx {1\over h_i}.
\end{equation}
The variance of the crosstalk noise on feature $i$ is also proportional to $1\over h_i$ accordingly. To see how the representation adapts, we expand $\mathcal L_{{\rm rep } }$ around the uniform solution $\mathcal L_{{\rm rep } }^{(0)}$ to first order in $\delta p_i\equiv p_i -\bar E/n$, treating $s_i, b_i$ and $h_i$ as independent variables. For a feature with $\delta p_i > 0$, we find $\delta h_i > 0$, $\delta s_i < 0$ and $\delta b_i > 0$ (Fig. \ref{fig:non-uniformity-loss}b). Therefore, a more frequent feature occupies more dimensions, receives less crosstalk noise, hence needs less filtering by its norm and bias. The leading order correction of $\mathcal L_{{\rm rep } }$ is
\begin{equation}
  \mathcal L_{{\rm rep } } = \mathcal L_{{\rm rep } }^{(0)} - c \sum_{i=1}^r \delta p_i^2
\end{equation}
with positive $c$. Therefore, the perturbation analysis captures the trends of $s_i$ and $b_i$, as well as the downward shift of $\mathcal L$, though it is quantitative only for weak non-uniformity.

\section{Related works}

The idea that networks represent more features than dimensions arose in studies of word embeddings \citep{gabriel_goh_decoding_2016,arora_linear_2018}. \citet{olah_zoom_2020} developed it as superposition hypothesis, which was demonstrated in the toy model of \citet{elhage_toy_2022}. In small models, the optimal representation vectors form polytopes \citep{elhage_toy_2022,chen_dynamical_2023}. Per-feature capacity has been quantified by the feature dimensionality of \citet{elhage_toy_2022}, refined into a constrained-allocation framework \citep{scherlis_polysemanticity_2025}, and developed into a spectral theory \citep{ivanov_spectral_2026}. In the large-system limit, \citet{cowsik_persian_2024} derived the loss by assuming a permutation-symmetric solution.

Loss decreases as a power law in model size across domains \citep{hestness_deep_2017,kaplan_scaling_2020,henighan_scaling_2020,hoffmann_training_2022}.The connection between neural scaling law and superposition was pointed out in \citet{liu_superposition_2025}, which gave the inverse width scaling supported in LLM experiments \citep{liu_neural_2026,liu_inverse_2026}. Our work generalizes their theory. 

Sparse coding represents data as sparse combinations of atoms from an overcomplete dictionary \citep{olshausen_emergence_1996}; dictionary learning recovers a ground-truth dictionary from such data \citep{spielman_exact_2012,agarwal_learning_2014,arora_simple_2015}, and sparse autoencoders apply it to network activations \citep{bricken_towards_2023,templeton_scaling_2024}, inverting the map that the toy model describes. When the dictionary is given, recovering the code is compressed sensing \citep{candes_robust_2006,donoho_compressed_2006}, where $\ell_1$ recovery fails sharply below $\sim k\log(n/k)$ measurements for $k$-sparse signals \citep{donoho_neighborliness_2005,donoho_counting_2008,donoho_observed_2009,amelunxen_living_2014}. This bound has been used to estimate the capacity of superposition \citep{elhage_toy_2022,klindt_unifying_2026} and to posit a critical width \citep{sarkar_geometric_2026}.

\section{Discussion}

In this paper we focus on the transition from partial to full representation. It has a simple origin: the competition between representing and discarding a feature. Representing a feature results in crosstalk that grows with active feature number $\bar{E}$ and decays with model width $m$, while discarding it results in an error depending on its variance, independent of $m$. Their balance sets the critical width $m^*\sim\bar{E}\log(n/\bar{E})$, with the loss dominated by the discarding error below $m^*$ and by crosstalk above it. Under non-uniformity, rare features are represented last, which delays the transition, while the loss is lowered since more frequent features occupy more hidden dimensions.

There are several limitations in this paper. Our work bases fully on the linear representation hypothesis. We focus on overall loss behavior rather than the loss from a finite dataset. We use Gaussian distributions to approximate the crosstalk noise, which discards the fine structure of representation vectors and overestimates the tail that leaks through the ReLU. We treat correlations only in exchangeable form (appendix \ref{sec:det-exp-results-exch-corr-feat}) and non-uniformity only at moderate strength. Finally, whether our results extend to other architectures and losses remains to be verified.

Parts of our results are consistent with observations in LLMs. \citet{liu_superposition_2025} found inverse width scaling in open-source LLMs in strong superposition. \citet{sarkar_geometric_2026} found that the loss of narrow students distilled from an LLM saturates at a floor due to discarded features when the width falls below a critical value scaling as $\bar{E}\log(n/\bar{E})$. Sparse autoencoders would allow more of our predictions to be tested in LLMs, but this is beyond the scope of this paper. Conversely, since $m^*$ and the loss are set by the data statistics, our results open the possibility of measuring the number of features and other statistics of natural language indirectly. Conceptually, since our results rely on generic ingredients: sparse features, a width bottleneck, and nonlinearity, we expect similar transitions in other neural networks. In conclusion, our results provide a step toward understanding the mechanism underlying LLMs.
\subsection*{AI use statement}

In this work, we used generative AI tools to assist with the development and checking of mathematical derivations. The tools also provided suggestions that informed parts of the theoretical analysis. The authors independently re-derived and verified the resulting mathematical claims.

We also used generative AI tools to generate and modify code for numerical experiments and visualizations, proofread and rewrite manuscript text for readability, critique the manuscript, assist with minor brainstorming, and search for relevant literature. The visualization code generated plots from author-provided data and did not generate the underlying data or results.

All AI-assisted material retained in the paper was reviewed by the authors. The code was checked through inspection, tests, and experiments, and all cited sources identified with AI assistance were independently verified. We take responsibility for the final content of this work, including all text, mathematical claims, analyses, code, figures, and other artifacts produced with the assistance of generative AI.






\bibliography{iclr2027_conference}
\bibliographystyle{iclr2027_conference}

\appendix
\section{Theoretical analysis}

\subsection{Non-superposition loss}
\label{sec:non-superp-loss}

In this section, we consider the non-superposition configuration. When the model exhibits no superposition, only $m$ features are represented (Fig. \ref{fig:intro-model}a). Assume the first $m$ features are represented,
\begin{equation}
  \label{eq:non-superposition-frame}
  \boldsymbol W = (\boldsymbol I_m~ \boldsymbol 0_{m\times (n-m)}).
\end{equation}
Correspondingly, the first $m$ biases $b_1=b_2=...=b_m=0$, so that the first $m$ features contribute no loss. To minimize the loss, the remaining biases generally do not vanish. We denote them by
\begin{equation}
\label{eq:non-superposition-rest-biases}
\tilde {\boldsymbol b} = (b_{{m+1}}, b_{m+2},...,b_{n})^T.
\end{equation}
Under this configuration, the output can be analytically given,
\begin{align}
  \boldsymbol y=&\left [ \boldsymbol W^T\boldsymbol W\boldsymbol x + \boldsymbol b \right ]_+ \\
  =&\left [ \left (\begin{array}{cc} \boldsymbol I_m & 0_{m\times (n-m)} \\ 0_{(n-m)\times m} & 0_{(n-m)\times (n-m)} \end{array}\right ) \boldsymbol x + \binom{\boldsymbol 0}{\tilde {\boldsymbol b}} \right ]_+ \\
  =&(x_1~ x_2~...~x_m~~ [b_{m+1}]_+~~ [b_{m+2}]_+~~...~~[b_n]_+)^T.
\end{align}
Therefore, the loss is
\begin{align}
  \mathcal L_{{\rm NS } }=&{1\over n}\mathbb E_{\boldsymbol x} \| \boldsymbol y-\boldsymbol x \|^2 \\
  =&{1\over n}\int \mathcal D v_{m+1}\mathcal D v_{m+2}...\mathcal D v_{n}\sum_{\boldsymbol u}p(\boldsymbol u)\sum_{i=m+1}^n\left ( [b_i]_+ - u_iv_i \right )^2, \label{eq:non-superposition-inter-1}
\end{align}
where $\mathcal D x$ is the shorthand of ${\rm d} x~p(x)$. By defining the marginal firing probabilities of features
\begin{equation}
\label{eq:def-marginal-firing-probability}
p_i = \sum_{\boldsymbol u}p(\boldsymbol u) u_i,
\end{equation}
Eq. (\ref{eq:non-superposition-inter-1}) can be simplified as
\begin{align}
  \mathcal L_{{\rm NS } }=&{1\over n}\sum_{i=m+1}^n \int \mathcal D v_i \left ( p_i ([b_i]_+ - v_i)^2+(1-p_i)[b_i]_+^2 \right ) \\
  =&{1\over n}\sum_{i=m+1}^n\left ( [b_i]_+^2 - 2p_i [b_i]_+ \bar v + p_i (\bar v^2+\sigma _v^2) \right ).
\end{align}
The minimum is
\begin{equation}
  \label{eq:append-non-superposition-loss}
\mathcal L_{{\rm NS } }={1\over n}\sum_{{i=m+1}}^n \left ( -p_i^2\bar v^2+p_i(\bar v^2+\sigma _v^2) \right ).
\end{equation}
The corresponding optimal bias is
\begin{equation}
b_i = p_i\bar v, ~~~~~ {\rm for }~ m+1 \le  i .
\end{equation}

\subsubsection{Suboptimality of the non-superposition configuration}
\label{sec:subopt-non-superp}

Given its loss, it is natural to ask if the non-superposition configuration is the optimal configuration. To answer this question, we construct the following superposition configuration, which outperforms the non-superposition configuration: among $n$ features, $m+1$ of them are represented. Representation vectors of feature 1 to feature $m$ form an orthonormal basis, and that of feature $m+1$ forms an antipode with feature $m$ \citep{elhage_toy_2022}. In this configuration, the gain of representing one more feature is
\begin{equation}
  -p_{m+1}^2\bar v^2+p_{m+1}(\bar v^2+\sigma _v^2),
\end{equation}
while the increasing loss is
\begin{equation}
  p_{m,m+1}\mathbb E_{v_m,v_{m+1}} \left ( ([v_m-v_{m+1}]_+ -v_m)^2 + ([v_{m+1}-v_{m}]_+ -v_{m+1})^2 \right ),
\end{equation}
where $p_{m,m+1}$ is the probability that both feature $m$ and $m+1$ fire.

For the case we consider in the main text, the features are independent with small firing probability $\bar E/n\ll 1$, the gain is of order $\bar E/n$ while the increasing loss is of order $\bar E^2/n^2$. Therefore, the non-superposition configuration is always suboptimal.

\subsection{Equal-norm frame and random projection approximation for uniform features}
\label{sec:deta-deriv-part-RP}

In this section, we use the random projection approximation to estimate the loss for uniform features. By ``uniform'', we mean all features are exchangeable. They do not need to be independent. Based on the symmetry of features, we consider the representation configuration called equal-norm frame, where all $\boldsymbol w_i$ in the weight matrix $\boldsymbol W$ have equal norm $s$. This acts as an ideal full representation, as the participation ratio $r$ defined in Eq. (\ref{eq:def-simulation-r}) is precisely $n$. Furthermore, we let $b_i=b$ to be equal for all features as well.

Firstly, we assume that there are precisely $E$ active features in each input, instead of a random number. In this case, the equal-norm loss is
\begin{align}
\mathcal L_{{\rm EN } }(E,m,n;s,b) =& {1\over n} \mathbb E_{\boldsymbol x} \|[\boldsymbol W^T\boldsymbol W \boldsymbol x+\boldsymbol b]_+ -\boldsymbol x \|^2   \\
=&{1\over n \binom{n}{E}}\sum_{a_1,...,a_E}\int {\mathcal  D} v_{a_1}{\mathcal  D} v_{a_2}...{\mathcal  D} v_{a_E}\sum_{i=1}^n \left \{ \left [\sum_{j=1}^E(\boldsymbol W^T\boldsymbol W)_{ia_j}v_{a_j}+b_i\right ]_+-x_i\right \}^2 \\
 =& {1\over n \binom{n}{E}}\sum_{a_1,...,a_E}\int {\mathcal  D} v_{a_1}...{\mathcal  D} v_{a_E}\Bigg \{ \sum_{i=1}^E \left [ \left [s^2v_{a_i}+\sum_{j:j\ne i} s^2c_{a_ia_j}v_{a_j}+b\right ]_+-v_{a_i}\right ]^2 \notag\\
&~~~~~~~~~~~~~~~~~~~~~~~~~~~~~~~~~~~~~~~~~~~~~~+\sum_{i\notin\{a_k\}} \left [ \sum_{j=1}^Es^2c_{ia_j}v_{a_j}+b\right]_+^2 \Bigg \}. \label{eq:append-precise-rewrite-loss}
\end{align}
The summation $\sum_{a_1,...,a_E}$ runs over all combinations of $E$ different active features among $n$, and
\begin{equation}
  \label{eq:append-def-cij}
  c_{ij}\equiv {\boldsymbol w_i \cdot \boldsymbol w_j \over s^2}
\end{equation}
are shorthands of cosine similarities of representation vectors. $\mathcal D v$ indicates integration is weighted by the probability density. The total loss is divided into two parts, one of which is the contribution of all active features, and the other is that of all inactive features. We denote them by $\mathcal L_A$ and $\mathcal L_I$ respectively. The crosstalk noises in each part are $\sum_{j:j\ne i} s^2c_{a_ia_j}v_{a_j}$ and $\sum_{j=1}^Es^2c_{ia_j}v_{a_j}$ respectively.

We take the limit $1\ll E$. By using mean field approximation, crosstalk noises can be regarded as independent random variables drawn from one distribution:
\begin{align}
\label{eq:append-def-noise-distribution}
  \sum_{j:j\ne i} s^2c_{a_ia_j}v_{a_j}\approx s^2 \eta ,&~~~~ \sum_{j=1}^Es^2c_{ia_j}v_{a_j}\approx s^2 \eta ,\\
  \eta=\sum_{i=1}^E c_{(i)}v_{(i)}.
\end{align}
Under the mean field approximation, Eq. (\ref{eq:append-precise-rewrite-loss}) becomes
\begin{align}
\mathcal L_{{\rm EN } }(E,m,n;s,b) \approx &{1\over n \binom{n}{E}}\sum_{a_1,...,a_E}\int {\mathcal  D} v_{a_1}...{\mathcal  D} v_{a_E}\mathcal D \eta \Bigg \{ \sum_{i=1}^E \left [ \left [s^2v_{a_i}+s^2\eta +b\right ]_+-v_{a_i}\right ]^2 \notag\\
            &~~~~~~~~~~~~~~~~~~~~~~~~~~~~~~~~~~~~~~~~~~~+\sum_{i\notin\{a_k\}} \left [s^2\eta +b\right]_+^2 \Bigg \}\\
  =& {1\over n}\left \{ E\int \mathcal D v\mathcal D\eta \left [[s^2v+s^2\eta +b]_+-v  \right ]^2 +(n-E)\int \mathcal D\eta \left [s^2\eta +b  \right ]_+^2\right \} \label{eq:append-loss-after-mean-field}\\
  =&{E\over n}\mathcal L_A + {n-E\over n}\mathcal L_I.
\end{align}

In order to calculate the loss, we need to know the distribution of $c_{ij}|_{i\ne j}$, which is analytically intractable for the trained frame. To proceed, we use Gaussian distribution to approximate $\mathcal D_{\eta }$. The mean value of $\eta $ cannot contribute to the loss, since we can always replace $b$ by $b-s^2\langle \eta \rangle $ to cancel its effect. Therefore, we simply let $\langle \eta \rangle =0$. (This is also approximately true in the experiment. The typical experimental result is $\langle c_{ij}|_{i\ne j} \rangle $ reaches the lower bound $-{1\over n-1}$, which is far less than $1\over m$ in the limit $m\ll n$. Consequently, $\langle \eta \rangle $ is also very small compared with the standard deviation $\sigma _{\eta }$. See Sec. \ref{sec:two-constr-repr}.) And
\begin{align}
\label{eq:append-var-noise}
  \sigma _{\eta }^2 =& E \sigma _c^2 (\bar v^2+\sigma _v^2)\\
  \ge& E (\bar v^2+\sigma _v^2){n-m \over m (n-1)},
\end{align}
where $\sigma _c^2$ is the variance of $c_{ij}|_{i\ne j}$, which has a lower bound
\begin{equation}
  \label{eq:append-lower-bound-sigmaC}
  \sigma _c^2\ge {n-m\over m(n-1)}
\end{equation}
when $\langle c_{ij}|_{i\ne j}\rangle $ is ignored. This lower bound is achieved when representation vectors form a tight frame. If the distribution of $\eta $ is confined to be Gaussian, the minimal loss is achieved when the variance is minimal, so we let $\sigma _{\eta }^2$ be its lower bound. Therefore, we have approximation (\ref{eq:xtalk-noise-approx-distribution}):
\begin{equation}
\label{eq:append-gaussian-approx-of-noise}
\eta \sim  \mathcal N\left (0, E (\bar v^2+\sigma _v^2){n-m \over m (n-1)}\right ).
\end{equation}
In other words, we assume that the trained representation approximately achieves a tight frame. With Eq. (\ref{eq:append-gaussian-approx-of-noise}), $\mathcal L_I$ can be further written as
\begin{equation}
\label{eq:append-loss-inactive}
\mathcal L_I = F\left ( -{b \over \sqrt{ 2s^4\sigma _{\eta }^2 }}\right ) s^4 \sigma _{\eta }^2,
\end{equation}
where
\begin{equation}
\label{eq:append-def-F}
F(\beta )\equiv  {2\beta ^2+1\over 2}(1-{\rm erf}\beta )-{\beta \over \sqrt{ \pi  } }\exp(-\beta ^2).
\end{equation}
Eq. (\ref{eq:append-loss-after-mean-field}) and Eq. (\ref{eq:append-gaussian-approx-of-noise}) give a complete estimation of the loss of the equal-norm frame.

In the limit $m\ll n$, the Gaussian approximation can also be interpreted as using random projection frame to approximate the trained optimal equal-norm frame. This is because in this limit,
\begin{equation}
\label{eq:append-gaussian-approx-vs-RP}
\sigma _{\eta }^2= E (\bar v^2+\sigma _v^2){n-m \over m (n-1)}\approx  {E\over m} (\bar v^2+\sigma _v^2).
\end{equation}
The Gaussian distribution with variance (\ref{eq:append-gaussian-approx-vs-RP}) is the distribution of $c_{ij}|_{i\ne j}$ in the random projection frame (this is also verified experimentally.) So we call mean field + Gaussian approximation \textbf{random projection (RP) approximation}.

In most cases, the number of active features in an input is not fixed. For example, in the setup of uniform independent features used in Sec. \ref{sec:trans-from-part-2-full} and \ref{sec:spars-isotr-indep}, the probability that $E$ active features occur is
\begin{equation}
\label{eq:ind-bernoulli-E-prob}
p_E = \binom{n}{E} \left (\bar E/ n\right )^E\left (1-\bar E/ n\right )^{n-E}.
\end{equation}
With variable $E$, the random projection approximation loss is simply the weighted average of losses:
\begin{align}
\label{eq:append-ind-binom-RP-loss}
  \mathcal L_{{\rm RP } }(\{p_E\}, m, n)=\min_{s,b} \sum_{E=0}^{\infty } p_E\Bigg \{& {n-E\over n}F\left ( -{b \over \sqrt{ 2 s^4 \sigma _{\eta }^2(E) }}\right ) s^4 \sigma _{\eta }^2(E) \notag \\
  &+{E\over n}\mathbb E_{v,\eta (E)} \left [ [s^2v+s^2\eta +b]_+-v  \right ]^2 \Bigg \}.
\end{align}
Eq. (\ref{eq:append-ind-binom-RP-loss}) is used numerically to estimate equal-norm frame loss without training. Since $\{p_E\}$ is determined by $\bar E$ for uniform independent features, we write $\bar E$ in that case.

\subsection{Estimating partial representation loss with partial RP approximation}
\label{sec:estim-part-repr-w-part-RP-approx}

In this section, we still consider uniform features. Based on the experimental result in Sec. \ref{sec:trans-from-part-2-full}, the partial representation is modeled as a partial equal-norm frame: $r$ features are represented by an equal-norm frame, and other $n-r$ features are just not represented, with $s_i=0$. Under this approximation, configurations with $r<m$ always have higher loss than that with $r=m$, hence we confine $r\in [m,n]$. The loss can be written as Eq. (\ref{eq:loss-total}):
\begin{equation}
\label{eq:loss-partial-representation}
\mathcal L(\{p_E\},m,n;r)= {r\over n}\mathcal L_{{\rm rep } }+{n-r\over n}\mathcal L_{{\rm unrep } },
\end{equation}
where $\mathcal L_{\rm  rep} $ is the loss contributed by represented features, and $\mathcal L_{{\rm unrep } }$ is that from features not represented. As in Sec. \ref{sec:deta-deriv-part-RP}, we use RP approximation (\ref{eq:append-ind-binom-RP-loss}) to estimate $\mathcal L_{{\rm rep } }$:
\begin{equation}
\label{eq:append-loss-represented-RP}
\mathcal L_{{\rm rep } }=\mathcal L_{{\rm RP } }\left ( \{p_{E_r}\}, m,r \right ),
\end{equation}
where $p_{E_r}$ is the probability of $E_r$ active features occurs among $r$ represented features. Eq. (\ref{eq:append-loss-represented-RP}) is just Eq. (\ref{eq:loss-rep}) in the main text. For uniform independent features,
\begin{equation}
  p_{E_r}=\binom{r}{E_r}\left (\bar E / n\right )^{E_r}\left (1-\bar E/n\right )^{r-E_r};
\end{equation}
for general uniform features,
\begin{equation}
  \label{eq:append-hypergeometric-pEr}
  p_{E_r}=\sum_{E=E_r}^n p_E \binom{r}{E_r}\binom{n-r}{E-E_r}\left / \binom{n}{E} \right . ,
\end{equation}
where $E$ is the total number of active features among all $n$ features.

$\mathcal L_{{\rm unrep } }$ can be written out as in Sec. \ref{sec:non-superp-loss}. Still, we assume last $(n-r)$ features are unrepresented:
\begin{align}
  \mathcal L_{\rm unrep}&={1\over n-r}\mathbb E_{\boldsymbol x}\sum_{i=r+1}^n(y_i-x_i)^2 \\
                        &={1\over n-r} \int \mathcal Dv_{r+1}...\mathcal Dv_{n}\sum_{\boldsymbol u}p(\boldsymbol u)\sum_{{i=r+1}}^n ([b_i]_+ -u_iv_i)^2 \\
  &={1\over n-r}\sum_{i=r+1}^n \int \mathcal D v_i \left ( p_i ([b_i]_+ - v_i)^2+(1-p_i)[b_i]_+^2 \right ) \\
                        &={1\over n-r}\sum_{i=r+1}^n\left ( [b_i]_+^2 - 2p_i [b_i]_+ \bar v + p_i (\bar v^2+\sigma _v^2) \right ) \\
  &\ge {1\over n-r}\sum_{i=r+1}^n \left ( -p_i^2\bar v^2+p_i(\bar v^2+\sigma _v^2) \right ), \label{eq:append-loss-not-represented}
\end{align}
where the marginal firing probabilities $p_i$ are defined in Eq. (\ref{eq:def-marginal-firing-probability}). The minimum is achieved with $b_i=p_i\bar v=\bar E \bar v /n$ for $r+1\le i$. In trained models, the biases of unrepresented features typically settle at negative values rather than at the optimum $p_i\bar v$, which is because for $b_i < 0$ the output is zero and the gradient on $b_i$ vanishes, while the resulting excess loss $p_i^2 \bar v^{2}$ is negligible.

Plugging Eq. (\ref{eq:append-loss-represented-RP}) and (\ref{eq:append-loss-not-represented}) into Eq. (\ref{eq:loss-partial-representation}), we obtain the \textbf{partial RP approximation}
\begin{equation}
\label{eq:append-partial-RP-loss}
\mathcal L_{{\rm pRP } }(\{p_E\},m,n;r)=  {r\over n}\mathcal L_{{\rm RP } }\left ( \{p_{E_r}\}, m,r \right )+{n-r\over n}\left (-{\bar E^2\over n^2}\bar v^2+{\bar E\over n}(\bar v^2+\sigma _v^2)\right ).
\end{equation}

Since in Eq. (\ref{eq:append-partial-RP-loss}), $r$ is not given but obtained in optimization, it can be used to calculate the transition point from partial representation to full representation. Following Eq. (\ref{eq:def-r-opt}),
\begin{align}
  \label{eq:append-def-opt-L}
  \mathcal L_{{\rm pRP } }(\{p_E\},m,n)&=\min_r\mathcal L_{{\rm pRP } }(\{p_E\},m,n;r),\\
  r_{{\rm opt } }(\{p_E\},m,n)&=\arg\min_r\mathcal L(\{ p_E \}, m, n; r).
\end{align}
The transition point $m^{*}$ is defined as
\begin{equation}
\label{eq:append-def-transition-m}
m^{*}\equiv  \inf_m\{ m:  r_{{\rm opt } }(\{p_E\},m,n) = n \}.
\end{equation}

\subsection{Asymptotic behaviors of partial RP approximation}
\label{sec:asympt-behav-part}

In this section, we analyse the asymptotic behaviors of partial RP approximation (\ref{eq:append-def-opt-L}) in different limits.

\subsubsection{Large width limit: $1\ll \bar E\ll m<n$}
\label{sec:large-width-limit}

\begin{figure}[htb]
\begin{center}
\includegraphics[width=0.5\textwidth]{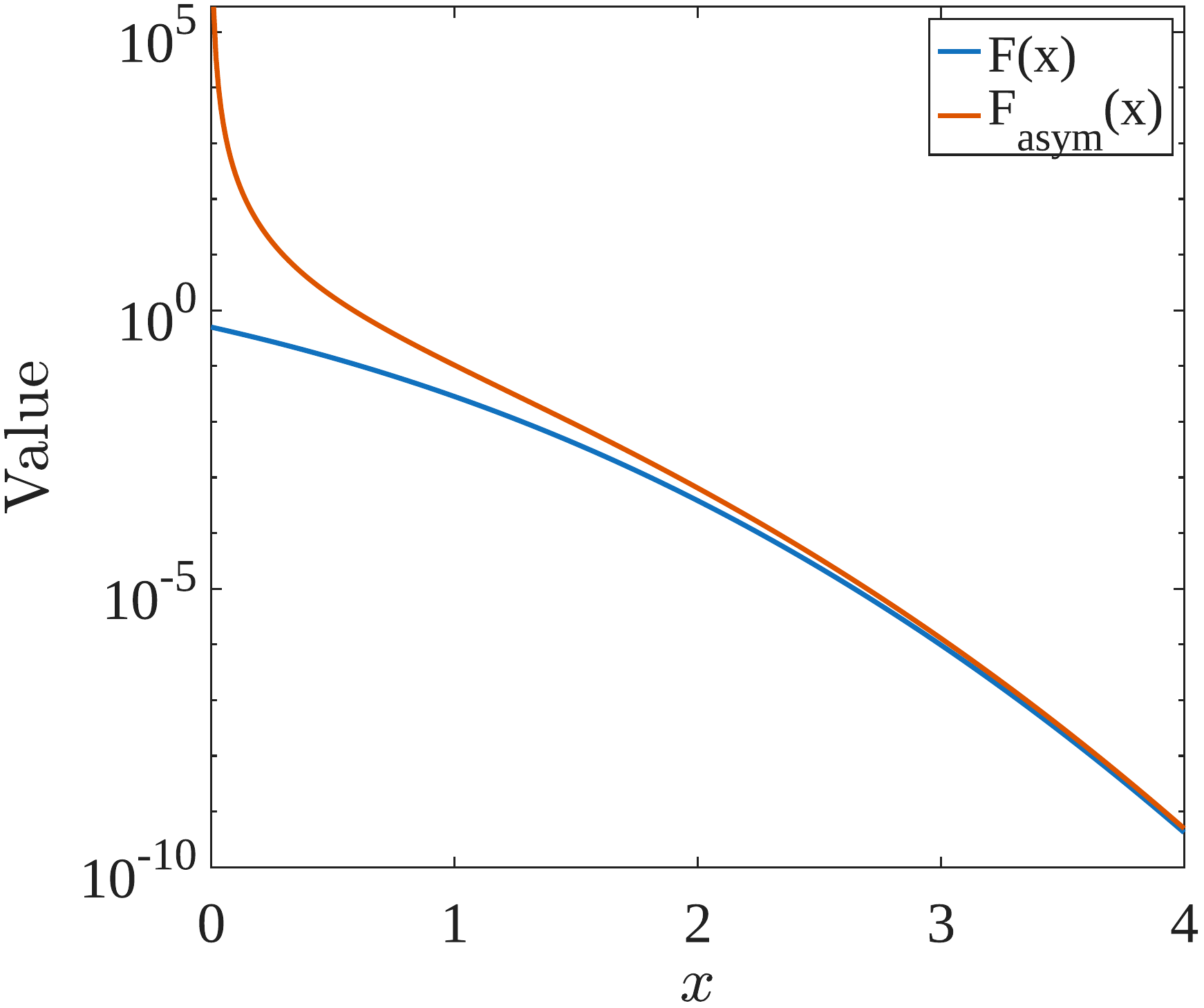}
\end{center}
\caption{Auxiliary function $F(x)$ and $F_{{\rm asym } }(x)$.}
\label{fig:append-FvsFasym}
\end{figure}

In this limit, typical crosstalk noise $\sigma _{\eta }^2\sim {\bar E r\over mn}{r-m \over r-1} (\bar v^2+\sigma _v^2) \ll 1 $. As a result, the bias to cancel the crosstalk noise on inactive represented features can also be small, typically $b\sim -\sqrt{ \sigma _{\eta }^2 }$, and to recover active represented features, the norm $s$ needs to be approximately 1. Therefore, active represented features are nearly not affected by ReLU, and in Eq. (\ref{eq:append-ind-binom-RP-loss})
\begin{align}
  \mathbb E_{v,\eta (E)}\left [ [s^2v+s^2\eta +b]_+ - v \right ]^2 &\approx \mathbb E_{{v,\eta (E)}}\left ( s^2v+s^2\eta +b-v \right )^2 \\
  &=(s^2-1)^2\sigma _v^2+[(s^2-1)\bar v+b]^2+s^4\sigma _{\eta }^2.
\end{align}
Based on the observation, we use the following ansatz for optimal $b$ and $s$:
\begin{align}
  b&=-\beta \sigma _{\eta }(\bar E) +O(\sigma _{\eta }^2(\bar E)), \\
  s&=1+\Delta s \sigma _{\eta }(\bar E) +O(\sigma _{\eta }^2(\bar E)).
\end{align}
Under this ansatz, the represented loss becomes
\begin{align}
  \mathcal L_{{\rm rep } }&=\mathcal L_{{\rm RP } }(\{p_{E_r}\},m,r) \notag\\
                          &\approx \sum_{E_r} p_{E_r} \left \{ F\left (\beta \sigma _{\eta }(\bar E) \over \sqrt{ 2 } \sigma_{\eta }(E_r)  \right ){\sigma _{\eta }^2(E_r) \over \sigma _{\eta }^2(\bar E) }+ {E_r\over r}\left [ 4\Delta s^2  \sigma _v^2+(2\Delta s\bar v-\beta )^2+{\sigma _{\eta }^2(E_r) \over \sigma _{\eta }^2(\bar E)} \right ]  \right \}\sigma _{\eta }^2(\bar E) \notag \\
                          &=  \sum_{E_r} p_{E_r} \left \{ F\left ({ \beta  \over \sqrt{ 2 } }\sqrt{ \bar E /E_r  } \right ){E_r \over \bar E }+ {E_r\over r}\left [ 4\Delta s^2  \sigma _v^2+(2\Delta s\bar v-\beta )^2+{E_r \over \bar E} \right ]  \right \}\sigma _{\eta }^2(\bar E) \notag \\
  &= \sum_{E_r} p_{E_r}  F\left ({ \beta  \over \sqrt{ 2 } }\sqrt{ \bar E /E_r  } \right ){E_r \over \bar E } \sigma _{\eta }^2(\bar E) +{\bar E \over n} \left [ 4\Delta s^2  \sigma _v^2+(2\Delta s\bar v-\beta )^2+{r\over n}+{r\sigma _{E_r}^2\over n\bar E_r^2} \right ]  \sigma _{\eta }^2(\bar E) \label{eq:append-leading-term-L-rep}
\end{align}
to the leading order. Here we also use $E_r/r \sim \bar E/n \ll 1$, and rewrite $\bar E_r$ as $\bar E r/n$ based on Eq. (\ref{eq:append-hypergeometric-pEr}). Eq. (\ref{eq:append-leading-term-L-rep}) is quadratic in $\Delta s$, so we can analytically optimize it:
\begin{equation}
  \mathcal L_{{\rm rep } }= \sum_{E_r} p_{E_r}  F\left ({ \beta  \over \sqrt{ 2 } }\sqrt{ \bar E /E_r  } \right ){E_r \over \bar E } \sigma _{\eta }^2(\bar E) +{\bar E \over n} \left [ {\beta ^2\sigma _v^2 \over \sigma _v^2+\bar v^2}+{r\over n}+{r\sigma _{E_r}^2\over n\bar E_r^2} \right ]  \sigma _{\eta }^2(\bar E), \label{eq:append-temp-expression-Lrep-w-beta}
\end{equation}
with
\begin{equation}
  \Delta s_{{\rm opt } }={\beta \bar v \over 2(\sigma _v^2+\bar v^2)}.
\end{equation}
The parameter $\beta $ can be optimized numerically. The auxiliary function $F$ decays exponentially (Fig. \ref{fig:append-FvsFasym}), hence the first term in Eq. (\ref{eq:append-temp-expression-Lrep-w-beta}) is of the same order of the second term at optimal $\beta $, even though $\bar E\ll n$. Intuitively, the ReLU + bias nonlinearity can filter most of crosstalk noise on inactive represented features. Therefore, at optimum, $\mathcal L_{{\rm rep } } \sim {\bar E \over n}\sigma _{\eta }^2(\bar E)$, which is far less than $\mathcal L_{{\rm unrep } }\sim {\bar E \over n}$ in Eq. (\ref{eq:append-loss-not-represented}). In this limit $1\ll \bar E \ll m <n$, all features are represented at optimum, i.e., $r_{{\rm opt } }=n$. The total loss to the leading order of $\bar E/m$ is
\begin{equation}
  \label{eq:append-LRP-var-E}
\mathcal L_{{\rm pRP } }(\{p_E\},m,n)=\min_{\beta }\left \{ \sum_E p_E  F\left ({ \beta  \over \sqrt{ 2 } }\sqrt{ \bar E /E  } \right ){E \over \bar E }  +{\bar E \over n} \left [ {\beta ^2\sigma _v^2 \over \sigma _v^2+\bar v^2}+1+{\sigma _{E}^2\over \bar E^2} \right ]  \right \} \sigma _{\eta }^2(\bar E).
\end{equation}
Eq. (\ref{eq:append-LRP-var-E}) shows that in this limit, the partial RP loss $\mathcal L_{{\rm pRP } }$ is proportional to $\sigma _{\eta }^2(\bar E)$. If we further apply the limit $m\ll n$,
\begin{equation}
  \mathcal L_{{\rm pRP } }(\{ p_E\},m,n)\propto \sigma _{\eta }^2(\bar E) \propto {1\over m}.
\end{equation}

Furthermore, we consider the limit $\sigma _E\ll \bar E$ for simplicity, which works for uniform independent features in the limit $1\ll \bar E$, as $\sigma _E=\sqrt{ \bar E (1-\bar E/n) }\ll \bar E$. Because of the highly peaked distribution of $E$, Eq. (\ref{eq:append-LRP-var-E}) can be approximated as
\begin{equation}
  \label{eq:append-LpRP-no-sigmaE-small-noise}
  \mathcal L_{{\rm pRP } }(\{p_E\},m,n)=\min_{\beta }\left \{  F\left ({ \beta  \over \sqrt{ 2 } } \right )  +{\bar E \over n} \left [ {\beta ^2\sigma _v^2 \over \sigma _v^2+\bar v^2}+1 \right ]  \right \} \sigma _{\eta }^2(\bar E).
\end{equation}
The optimal $\beta$ is achieved when
\begin{equation}
  \label{eq:append-optimal-beta-condition}
  {1\over \sqrt{ 2 } } F'(\beta /\sqrt{ 2 } ) + {\bar E\over n}{2\sigma _v^2 \beta  \over \sigma _v^2+\bar v^2}=0.
\end{equation}
The optimal $\beta $ grows with $n/\bar E$. In the large $\beta $ limit, the following asymptotic equation holds true (Fig. \ref{fig:append-FvsFasym}):
\begin{equation}
  \label{eq:append-asymp-F}
  F(x) =F_{{\rm asym } }(x)\equiv  {e^{-x ^2}\over 2\sqrt{ \pi  } x ^3 }~~~~~{\rm as~ }x\to \infty  .
\end{equation}
The asymptotic solution of Eq. (\ref{eq:append-optimal-beta-condition}) is
\begin{equation}
  \beta =\sqrt{ 2 \log {n\over \bar E} }.
\end{equation}
The asymptotic expression of $\mathcal L_{{\rm pRP } }$ becomes
\begin{equation}
  \label{eq:append-LpRP-lim-1-ll-E-ll-m-n}
  \mathcal L_{{\rm pRP } }=2\sigma _v^{2}{\bar E^2\over nm}{n-m\over n-1}\log{n\over \bar E}.
\end{equation}
In the limit $1\ll \bar E\ll m \ll n$, this expression degenerates to Eq. (\ref{eq:asymptotic-total-loss}):
\begin{equation}
  \label{eq:append-LpRP-lim-1-ll-E-ll-m-ll-n}
  \mathcal L_{{\rm pRP } }={2\sigma _v^2\bar E^2 \over nm} \log{n\over \bar E}.
\end{equation}

For partial representation configurations, the derivation of Eq. (\ref{eq:append-LpRP-lim-1-ll-E-ll-m-n}) holds true for represented features in the limit $1\ll r\bar E/n \ll m < r$ and $\sigma _{E_r}\ll \bar E_r=r\bar E/n$, and the approximate $\mathcal L_{{\rm rep } }$ is
\begin{equation}
  \mathcal L_{{\rm rep } }=2\sigma _v^2{\bar E^2 r\over n^2m}{r-m\over r-1}\log{n\over \bar E}.
\end{equation}
If we further assume $m\ll r$,
\begin{equation}
\label{eq:append-Lrep-lim-1-ll-Er-ll-m-r}
  \mathcal L_{{\rm rep } }={2\sigma _v^2\bar E^2r \over n^2m}\log {n\over \bar E}.
\end{equation}

Based on  Eq. (\ref{eq:append-Lrep-lim-1-ll-Er-ll-m-r}), we can estimate the critical width $m^{*}$. By plugging Eq. (\ref{eq:append-Lrep-lim-1-ll-Er-ll-m-r}) in Eq. (\ref{eq:append-partial-RP-loss}), the partial RP loss can be written as
\begin{equation}
  \label{eq:append-temp-expression-LpRP}
  \mathcal L_{{\rm pRP } }(\bar E,m,n;r)={2\sigma _v^2r\over n}{\bar E^2 r\over n^2m}\log{n\over \bar E}+{n-r\over n}{\bar E\over n}(\bar v^2+\sigma _v^2).
\end{equation}
According to Eq. (\ref{eq:append-temp-expression-LpRP}), the optimal $r$ is
\begin{equation}
  \label{eq:append-ropt-lim-1-ll-Er-ll-m-r}
  r_{{\rm opt } }=\min\left ({mn(\bar v^2+\sigma _v^2)\over 4\bar E\sigma _v^2\log{n\over \bar E}}, n\right ).
\end{equation}
The optimal $r_{{\rm opt } }$ reaches $n$ at $m^{*}$, hence
\begin{equation}
  \label{eq:append-m-trans-lim-1-ll-E-ll-n}
  m^{*}={4\sigma _v^2\bar E \over (\bar v^2+\sigma _v^2) } \log {n\over \bar E}.
\end{equation}
Note that $m^{*}$ is twice the width at which representing a feature and sacrificing it cost the same (the heuristic argument in Sec. \ref{sec:spars-isotr-indep}). This is because increasing $r$ from $n-1$ to $n$ not only costs $\mathcal L_{{\rm rep } }$, but also increases the loss of all other represented features, which results in a factor 2. Eq. (\ref{eq:append-ropt-lim-1-ll-Er-ll-m-r}) also predicts a constant $r_{{\rm opt } }\over m$ in partial representations, which requires $\log {n\over \bar E}\gg 1$ and cannot be implemented in experiment. Also, Eq. (\ref{eq:append-m-trans-lim-1-ll-E-ll-n}) cannot capture the trend of $m^{*}$ at different $\sigma _v$, see Sec. \ref{sec:effect-sigma-_v}.

\subsubsection{Weak superposition limit: $r- m\ll m$}
\label{sec:weak-superp-limit}

In Sec. \ref{sec:subopt-non-superp} we show the non-superposition configuration is suboptimal for uniform independent features with small firing probability. Here we show that the partial RP approximation gives the same result. We maintain the limit condition $1\ll \bar E \ll n$ and $\sigma _E\ll \bar E$. For partial RP approximation we additionally require $\bar E_r=m\bar E/n\gg 1$, i.e., $m\gg n/\bar E$. We let
\begin{equation}
  r=m+\Delta r,~~~{\rm where~ } \Delta r\ll m.
\end{equation}
The corresponding crosstalk noise variance is
\begin{equation}
  \sigma _{\eta }^2(\bar E_r)={\bar E r\over n}(\bar v^2 + \sigma _v^2) {r-m\over m(r-1)}\approx {\bar E \Delta r\over nm}(\bar v^2 + \sigma _v^2)\ll 1.
\end{equation}
Since typical crosstalk noise is small, derivation similar to Sec. \ref{sec:large-width-limit} can be applied here. Here we define
\begin{align}
  b&=-\beta \sigma _{\eta }(\bar E_r)+{\rm h.o.t. } ,\\
  s&=1+\Delta s \sigma _{\eta }(\bar E_r)+{\rm h.o.t. } 
\end{align}
The represented loss is
\begin{align}
  \mathcal L_{{\rm rep } }( r)\approx&\min_{\beta ,\Delta s}\left \{ F(\beta / \sqrt{ 2  })+{\bar E\over n}\left [ 4\Delta s^2\sigma _v^2+(2\Delta s \bar v-\beta )^2+1 \right ]\right \}\sigma _{\eta }^2(\bar E_r) \\
  =&\min_{\beta } \left \{ F(\beta / \sqrt{ 2  })+{\bar E\over n}\left [ {\beta ^2 \sigma _v^2 \over \sigma _v^2+\bar v^2}+1 \right ]\right \}\sigma _{\eta }^2(\bar E_r) . \label{eq:append-auxiliary-Lrep-weak-sup}
\end{align}
As discussed in Sec. \ref{sec:large-width-limit}, the first factor in Eq. (\ref{eq:append-auxiliary-Lrep-weak-sup}) is of order $\bar E\over n$ at optimal $\beta $, hence the loss of representing extra $\Delta r$ features is
\begin{equation}
  r\mathcal L_{{\rm rep } }(r) \sim {\bar E^2\Delta r\over n^2 }.
\end{equation}
On the other hand, the gain of representing extra $\Delta r$ features is
\begin{equation}
  -\Delta r \mathcal L_{{\rm unrep } }\approx -\Delta r {\bar E\over n}.
\end{equation}
The gain is much greater than the loss, which indicates that the non-superposition configuration is never the optimal configuration.

\subsubsection{Lowest-order correction from the variability of $E$}
\label{sec:deta-deriv-loss-exch-corr-feat}

In Sec. \ref{sec:large-width-limit} and \ref{sec:weak-superp-limit}, we neglect the effect of variable $E$. Here we derive the lowest-order correction on the partial RP loss of the fluctuation of $E$, in the limit $\sigma _{E_r}\ll \bar E_r$. Since $\mathcal L_{{\rm unrep } }$ is only determined by the marginal firing probabilities $\bar E/n$, we only need to correct $\mathcal L_{{\rm rep } }$. In the limit $1\ll \bar E_r \ll m < r \le n $, with
\begin{align}
  b\approx &-\beta \sigma _{\eta }(\bar E_r), \\
  s\approx &1+\Delta s\sigma _{\eta }(\bar E_r).
\end{align}
we have
\begin{align}
  \mathcal L_{{\rm rep } }=&\mathcal L_{{\rm RP } }(\{p_{E_r}\}, m,r) \notag \\
  =&\min_{\beta }\left \{ \sum_{E_r} p_{E_r}  F\left ({ \beta  \over \sqrt{ 2 } }\sqrt{ \bar E_r \over E_r  } \right ){E_r \over \bar E_r }  +{\bar E \over n} \left [ {\beta ^2\sigma _v^2 \over \sigma _v^2+\bar v^2}+1+{\sigma _{E_r}^2\over \bar E_r^2} \right ] \right \}  \sigma _{\eta }^2(\bar E_{r}) 
\end{align}
By expanding $F$ around $\beta/\sqrt{ 2 }$, the represented loss $\mathcal L_{{\rm rep } }$ can be further simplified as
\begin{align}
  \label{eq:append-temp-Lrep-correction-sigmaE-opt-beta}
  \mathcal L_{{\rm rep } }=&\min_{\beta }\bigg \{ F\left (\beta \over \sqrt{ 2 }\right )-F'\left (\beta \over \sqrt{ 2 }\right ){\beta \sigma _{E_r}^2 \over 8\sqrt{ 2  } \bar E_r^2} +F''\left ({\beta \over \sqrt{ 2 } }\right ) {\beta ^2 \sigma _{E_r}^2 \over 16 \bar E_r^2}  \notag\\
  &~~~~~~~~~~+{\bar E \over n} \left [ {\beta ^2\sigma _v^2 \over \sigma _v^2+\bar v^2}+1+{\sigma _{E_r}^2\over \bar E_r^2} \right ] \bigg \}  \sigma _{\eta }^2(\bar E_{r}) .
\end{align}
Since $F'$ is negative and $F''$ is positive, the correction increases $\mathcal L_{{\rm rep } }$. The optimal $\beta $ at $\sigma _{E_r}^2=0$ is denoted by $\beta _0$,
\begin{equation}
  \label{eq:append-beta-correction-sigmaE}
  \beta =\beta _0 + \Delta \beta \sigma _{E_r}^2 + O(\sigma _{E_r}^4),
\end{equation}
with
\begin{equation}
\Delta \beta \equiv {F'(\beta_0/\sqrt{ 2 }){1\over 8\sqrt{ 2 }} -  F''(\beta _0/\sqrt{ 2 }){\beta _0\over 16} + F'''(\beta _0/\sqrt{ 2 }) {\beta _0^3 \over 16\sqrt{ 2 } } \over {1\over 2} F''(\beta _0/\sqrt{ 2 }) + {2\bar E\over n}{\sigma _v^2 \over \sigma _v^2+\bar v^2} }{1 \over \bar E_r^2}.
\end{equation}
Plugging Eq. (\ref{eq:append-beta-correction-sigmaE}) into Eq. (\ref{eq:append-temp-Lrep-correction-sigmaE-opt-beta}), the loss $\mathcal L_{{\rm rep } }$ grows linearly with $\sigma _E^2$ to the lowest order. Therefore, we expect $m^{*}$ to increase with $\sigma _E^2$, as increased $\mathcal L_{{\rm rep } }$ requires larger width to outperform $\mathcal L_{{\rm unrep } }$.

\subsection{Perturbation analysis for independent features with unequal firing probabilities}
\label{sec:pert-analys-indep}

In this section, we consider the independent features with weakly non-uniform firing probabilities:
\begin{equation}
  p_i = p_0+\delta p_i,~~~{\rm for~ } i=1,...,n,
\end{equation}
where $p_0$ is the mean firing probability, and the deviation $\delta p_i$ is far less than $p_0$. Since the non-uniformity is weak, we can treat its effect perturbatively. In Sec. \ref{sec:leverage-score-as}, we introduce the leverage score, which we use to quantify the space each feature occupies. Then in Sec. \ref{sec:lowest-order-pert-to-part-RP}, we derive the lowest-order perturbative correction to the partial RP approximation.

\subsubsection{The leverage score as an effective dimensionality}
\label{sec:leverage-score-as}

We define the leverage score for represented feature and set the leverage score of zero vectors to be 0 by convention. For simplicity, here we assume all features are represented. Let $\tilde {\boldsymbol W}$ be the normalized frame
\begin{equation}
  \label{eq:append-normalized-frame}
  \tilde {\boldsymbol W}=\left ({\boldsymbol w_1 \over s_1} ~ {\boldsymbol w_2 \over s_2} ~ ... ~ {\boldsymbol w_n \over s_n}\right ).
\end{equation}
As in Eq. (\ref{eq:def-leverage-score}), the leverage score $h_i$ of feature $i$ is defined as
\begin{equation}
  \label{eq:append-def-leverage-score}
  h_i\equiv {1\over s_i^2}\boldsymbol w_i^T(\tilde {\boldsymbol W}\tilde {\boldsymbol W}^T)^+{\boldsymbol w_i}.
\end{equation}
In Eq. (\ref{eq:append-def-leverage-score}) the operator $(\cdot )^+$ indicates the Moore-Penrose inverse; when $\tilde {\boldsymbol W}\tilde {\boldsymbol W}^T$ is invertible, it is simply the inverse as in Eq. (\ref{eq:def-leverage-score}). The leverage score can be interpreted as an effective dimensionality that a feature occupies. It possesses the following properties:
\begin{itemize}
\item Normalization: If vectors span $\mathbb  R^m$,
  \begin{equation}
    \label{eq:append-leverage-score-normalization}
    \sum_ih_i=m.
  \end{equation}
\item Invariance under global rotation: If $\boldsymbol W$ is replaced by $\boldsymbol R\boldsymbol W$ (where $\boldsymbol R \in O(m)$), the leverage score $h_i$ of each feature remains unchanged.
\item Divisibility: if among all vectors, $k$ vectors (subset $A$) span a $d$-dimensional subspace, and all other vectors (subset $B$) are perpendicular to this subspace. Then the leverage score of features in $A$ satisfies $\sum_{i\in A}h_i=d$.
\item Reasonable range: $h_i\in [0,1]$. A vectors at most occupies one dimension, which happens when all other vectors are orthogonal to it. And the leverage score of a zero vector is zero by convention.
\end{itemize}

We use $h_i$ to estimate the overlap between the representation vector of feature $i$ and other vectors in the limit of weak non-uniformity. The Gram matrix and frame operator of $\tilde {\boldsymbol W}$ are denoted respectively by
\begin{align}
\boldsymbol G=\tilde {\boldsymbol W} ^T\tilde {\boldsymbol W} \in \mathbb R^{n\times n},\\
\boldsymbol S=\tilde {\boldsymbol W} \tilde {\boldsymbol W} ^T\in \mathbb R^{m\times m}.
\end{align}
The elements of the Gram matrix $G_{ij}=\boldsymbol w _i\cdot \boldsymbol w _j/s_is_j$ are cosine similarities of vectors, satisfying
\begin{align}
\sum_{j:j\ne i}\left ({\boldsymbol w _i\cdot \boldsymbol w _j \over  s_is_j}\right )^2=&\sum_{j}G_{ij}^2-G_{ii}^2\\
=&(\boldsymbol G\boldsymbol G^T)_{ii}-1.
\end{align}
Let
\begin{equation}
  \tilde {\boldsymbol W}=\boldsymbol U \boldsymbol \Sigma \boldsymbol V^T
\end{equation}
be the SVD decomposition of $\tilde {\boldsymbol W}$. Therefore,
\begin{align}
  \sum_{j:j\ne i}\left ({\boldsymbol w _i\cdot \boldsymbol w _j \over  s_is_j}\right )^2=&(\boldsymbol G\boldsymbol G^T)_{ii}-1 \notag\\
=&(\tilde {\boldsymbol W}^T \tilde {\boldsymbol W}  \tilde {\boldsymbol W} ^T \tilde {\boldsymbol W} )_{ii}-1\\
=&(\boldsymbol  V\boldsymbol \Sigma ^T\boldsymbol U^T~\boldsymbol U\boldsymbol \Sigma \boldsymbol V^T~\boldsymbol  V\boldsymbol \Sigma ^T\boldsymbol U^T~\boldsymbol U\boldsymbol \Sigma\boldsymbol  V^T)_{ii}-1\\
=&\sum_{j:j\le m}V_{ij}^2\sigma _j^4-1, \label{eq:append-cos-sim-temp}
\end{align}
where  $\sigma _j$ are singular values of $\tilde {\boldsymbol W}$. Meanwhile,
\begin{align}
h_i=&[\tilde {\boldsymbol W}^T(\tilde {\boldsymbol W}\tilde {\boldsymbol W}^T)^+\tilde {\boldsymbol W}]_{ii}\\
=& \left (\boldsymbol V\boldsymbol \Sigma ^T\boldsymbol U^T(\boldsymbol U\boldsymbol \Sigma \boldsymbol V^T~\boldsymbol  V\boldsymbol \Sigma ^T\boldsymbol U^T)^+\boldsymbol U\boldsymbol \Sigma\boldsymbol  V^T\right )_{ii}\\
=&\sum_{j:j\le m}V_{ij}^2. \label{eq:append-leverage-score-temp}
\end{align}
By comparing Eq. (\ref{eq:append-cos-sim-temp}) and Eq. (\ref{eq:append-leverage-score-temp}), we have
\begin{align}
  \sum_{j:j\ne i}\left ({\boldsymbol w _i\cdot \boldsymbol w _j \over  s_is_j}\right )^2&=h_i{\sum_{j:j\le m} V_{ij}^2\sigma _j^4\over \sum_{j:j\le m} V_{ij}^2}-1 \notag \\
  &=h_i\left [\left ({\rm Mean }_i^V\sigma_j ^2 \right )^2+{\rm Var }_i^V\sigma _j^2\right ]  -1, \label{eq:append-accurate-crosstalk-overlap}
\end{align}
where
\begin{align}
  {\rm Mean }_i^V\sigma_j ^2&\equiv {\sum_{j:j\le m} V_{ij}^2\sigma _j^2\over \sum_{j:j\le m} V_{ij}^2} ={G_{ii}\over h_i} ={1\over h_i}, \\
  {\rm Var  }_i^V\sigma_j ^2&\equiv {\sum_{j:j\le m} V_{ij}^2(\sigma _j^2-{\rm Mean }_i^V\sigma_j ^2)^2\over \sum_{j:j\le m} V_{ij}^2}.
\end{align}
For tight frames (which we assume the trained representations approximately achieve to minimize $\sigma _{\eta }^2$, see Sec. \ref{sec:deta-deriv-part-RP}), all $\sigma _j$ are equal and the variance ${\rm Var  }_i^V\sigma_j ^2$ vanishes. Following it, we assume the optimal representation is nearly tight when the non-uniformity is weak, i.e.,  ${\rm Var  }_i^V\sigma_j ^2\ll {1\over h_i^2}\sim {n^2\over m^2}$. Furthermore, we assume $m\ll n$. Therefore,
\begin{equation}
  \label{eq:append-bridge-xtalk-leverage-score}
  \sum_{j:j\ne i}\left ( \boldsymbol w_i \cdot \boldsymbol w_j \over s_i s_j \right )^2\approx{1\over h_i}-1 \approx  {1\over h_i}.
\end{equation}
When $h_i={m\over n}$ are all equal, Eq. (\ref{eq:append-bridge-xtalk-leverage-score}) is consistent with Eq. (\ref{eq:append-lower-bound-sigmaC}).

Based on Eq. (\ref{eq:append-bridge-xtalk-leverage-score}), the crosstalk noise on feature $i$ is approximated as a Gaussian random variable with mean zero and variance
\begin{equation}
  \sigma _{s^2\eta , i}^2 \equiv \mathbb E_{\boldsymbol x}\left [s_i\sum_{j\in \{a_k\}\backslash i}v_js_j\left ( \boldsymbol w_i \cdot \boldsymbol w_j \over s_i s_j \right )\right ]^2 \approx  {s_i^2\langle  s^2 p\rangle  \over  h_i}(\bar v^2+\sigma _v^2) ,
  \label{eq:append-nonuni-xtalk-noise-variance-meanfield}
\end{equation}
where $\{a_k\}$ is the set of active features in an input, and $\langle s^2 p\rangle\equiv {1\over n}\sum_i s_i^2p_i$. Here we apply the mean field approximation on the crosstalk noise again.

\subsubsection{Lowest-order perturbative correction to the representation and the loss}
\label{sec:lowest-order-pert-to-part-RP}

Using the leverage score, we can estimate the loss at weak non-uniformity perturbatively. We first assume that all features are represented, so the random projection approximation (\ref{eq:append-ind-binom-RP-loss}) acts as the baseline of the perturbation analysis. We work in the limit $1\ll \bar E=p_0n\ll m\ll n$ and ignore the effect of $\sigma _E$. Similar to Eq. (\ref{eq:append-leading-term-L-rep}), the loss is approximated by
\begin{align}
  \mathcal L(\{p_i\},m,n;\{b_i\},\{s_i\},\{h_i\}) ={1\over n}\sum_{i=1}^n\mathcal L&_f(s_i,b_i,h_i,p_i,\langle s^2p\rangle ) \label{eq:append-nonuniform-loss-divided} \\
  ={1\over n} \sum_i\Bigg [&(1-p_i)F\left ( -{b_i \over \sqrt{ 2  \sigma _{s^2\eta , i}^2 } }  \right ) \sigma _{s^2\eta , i}^2 \notag \\
    &+p_i\left ( (s^2_i-1)^2\sigma _v^2+(s^2_i-1+b_i )^2+ \sigma _{s^2\eta , i}^2  \right )\Bigg ].
\end{align}
In Eq. (\ref{eq:append-nonuniform-loss-divided}), $\mathcal L_f$ is the loss contribution of a single feature. The predicted loss is obtained by optimizing $\{b_i\},\{s_i\}$ and $\{h_i\}$ under the constraint (\ref{eq:append-leverage-score-normalization}). Therefore, we use the method of Lagrange multipliers, and let
\begin{equation}
  \tilde {\mathcal L}\equiv n\mathcal L+A\left (\sum_i h_i-m\right ).
\end{equation}
Baseline parameters are
\begin{align}
h_0=&{m\over n},\\
(s_0,b_0)=&\arg\min_{(s,b)}\mathcal L_f(s,b,h_0,p_0,\langle s^2p\rangle)|_{\langle s^2p\rangle =s^2p_0}. \label{eq:append-nonuni-baseline-sb}
\end{align}
And we define
\begin{align}
  s_i&=s_0+\delta s_i,\\
  b_i&=b_0+\delta b_i,\\
  h_i&=h_0+\delta h_i.
\end{align}
We expand $\tilde {\mathcal L}$ around the baseline parameters:
\begin{align}
  &\tilde {\mathcal L}(\{s_i\},\{b_i\},\{h_i\},\{p_i\},A) \notag
  \\=&n {\mathcal L}|_0+\sum_i\Bigg ( {\partial \mathcal L_f\over \partial s}\bigg|_0 \delta s_i+ {\partial \mathcal L_f\over \partial b}\bigg|_0 \delta b_i+{\partial \mathcal L_f\over \partial p}\bigg|_0 \delta p_i + {\partial \mathcal L_f\over \partial h}\bigg|_0 \delta h_i\notag\\
  &~~~~~~~~~~~~~~~~~~~~~~+A\delta h_i + {\partial \mathcal L_f\over \partial \langle s^2p\rangle }\bigg|_0\cdot {1\over n}\sum_k\left (2p_0s_0 \delta s_k +s_0^2\delta p_k + p_0\delta s_k^2+ 2s_0\delta s_k\delta p_k \right ) \Bigg )\notag\\
&+\sum_i{1\over 2}(\delta s_i,\delta b_i,\delta h_i;\delta p_i)\cdot \mathcal H|_0\cdot (\delta s_i,\delta b_i,\delta h_i;\delta p_i)^T+{\rm h.o.t. },
\end{align}
where $(\cdot )|_0$ denotes the function evaluated at $s_0,b_0,h_0$ and $p_0$, and $\mathcal H$ is the Hessian matrix of $\mathcal L_f$ at fixed $\langle s^2p\rangle =s_0^2p_0$. First order terms vanish because of Eq. (\ref{eq:append-leverage-score-normalization}) and (\ref{eq:append-nonuni-baseline-sb}). So $\tilde {\mathcal L}$ can be simplified as
\begin{align}
  \tilde {\mathcal L}=&n\mathcal L|_0+\sum_i\left [{1\over 2}(\delta s_i,\delta b_i,\delta h_i;\delta p_i)\cdot \mathcal H|_0\cdot (\delta s_i,\delta b_i,\delta h_i;\delta p_i)^T + {\partial \mathcal L_f\over \partial \langle s^2p\rangle }\bigg|_0\left ( p_0\delta s_i^2+ 2s_0\delta s_i\delta p_i \right ) \right ]\\
  =&n\mathcal L|_0+{1\over 2}\sum_i\left ((\delta s_i,\delta b_i,\delta h_i)\cdot \mathcal H'|_0\cdot (\delta s_i,\delta b_i,\delta h_i)^T+2(\delta s_i,\delta b_i,\delta h_i)\cdot\mathcal V'|_0\delta p_i+\mathcal H_{pp}|_0\delta p_i^2\right ), \label{eq:append-loss-expansion-nonuni-HV}
\end{align}
where
\begin{equation}
  {\mathcal H}'\equiv \left (\begin{array}{ccc} \mathcal H_{ss} + 2p{\partial \mathcal L_f\over \partial \langle s^2p\rangle} & \mathcal H_{sb} & \mathcal H_{sh} \\\mathcal H_{bs} & \mathcal H_{bb} & \mathcal H_{bh} \\
 \mathcal H_{hs} & \mathcal H_{hb} & \mathcal H_{hh}\end{array}\right )\\,~~~~~~\mathcal V'\equiv \left (\begin{array}{ccc} \mathcal H_{sp}+2s{\partial \mathcal L_f\over \partial \langle s^2p\rangle } \\\mathcal H_{bp} \\ \mathcal H_{hp}\end{array}\right ).
\end{equation}
Therefore, the optimal parameters satisfy
\begin{equation}
  \label{eq:append-susceptibility-sbh}
  \left (\begin{array}{ccc} \delta s_i \\\delta b_i \\ \delta h_i\end{array}\right )=-(\mathcal H'|_0)^{-1}\cdot \mathcal V'|_0~\delta p_i.
\end{equation}
Correspondingly, the optimal loss $\mathcal L$ is
\begin{equation}
  \label{eq:append-correction-loss-nonuni}
\mathcal L(\{p_i\},m,n)=\mathcal L|_{0} -c\sum_i\delta p_i^2,
\end{equation}
with
\begin{equation}
  \label{eq:append-coeff-correction-loss-nonuni}
  c\equiv{1\over 2n}\left [ (\mathcal V'|_0)^T(\mathcal H'|_0)^{-1}\mathcal V'|_0-\mathcal H_{pp}|_0\right ].
\end{equation}
Eq. (\ref{eq:append-susceptibility-sbh}) and (\ref{eq:append-correction-loss-nonuni}) give the lowest-order correction to the representation and the loss.

For partial representation, the correction is applied to represented features, with $n$ replaced by $r$, $\langle s^2p\rangle $ replaced by the mean over represented features, $h_0$ replaced by $ m\over r$, and $p_{0}$ taken as the mean firing probability of represented features. 

\subsubsection{Discussion on the lowest-order perturbative correction}
\label{sec:disc-lowest-order}

We first discuss the signs of the corrections. In the experiments with uniform features, the norms and biases of all represented features are nearly equal at the loss minimum (Fig. \ref{fig:transition}ab). Therefore, at $\delta p_i=0$ the optimum is $\delta s_i=\delta b_i=\delta h_i=0$, i.e., the uniform solution is a local minimum. This ensures $\mathcal L$ in Eq.~(\ref{eq:append-loss-expansion-nonuni-HV}) increases with any fluctuation on $\delta s_i,\delta b_i$ and $\delta h_i$, i.e., $\mathcal H'|_0$ is positive definite. Moreover, at fixed $\langle s^2p\rangle$, $\mathcal L_f$ in Eq.~(\ref{eq:append-nonuniform-loss-divided}) is affine in $p_i$, hence $\mathcal H_{pp}|_0=0$. Following Eq. (\ref{eq:append-coeff-correction-loss-nonuni}),
\begin{equation}
    c=\frac{1}{2n}(\mathcal V'|_0)^T(\mathcal H'|_0)^{-1}\mathcal V'|_0>0
\end{equation}
for generic parameters, where $\mathcal V'|_0\neq 0$. Therefore, non-uniformity lowers the loss. The signs of the responses $\delta s_i$, $\delta b_i$ and $\delta h_i$ depend on the values of the Hessian elements, so we verify them numerically. We solve Eq.~(\ref{eq:append-susceptibility-sbh}) for all parameters we tested in toy model experiments. For all tested parameters, $\mathcal H'|_0$ is positive definite, and
\begin{equation}
  \label{eq:append-signs-of-susceptibilities}
    \frac{\delta h_i}{\delta p_i}>0,\quad \frac{\delta b_i}{\delta p_i}>0,\quad \frac{\delta s_i}{\delta p_i}<0.
\end{equation}
Therefore, a more frequent feature occupies more dimensions, receives less crosstalk noise, and hence needs less filtering by its norm and bias.

We emphasize that the perturbation analysis relies on approximations beyond the RP approximation, so it should be regarded as an ansatz. First, we neglect the term $h_i\mathrm{Var}^V_i\sigma_j^2$ in Eq.~(\ref{eq:append-accurate-crosstalk-overlap}). This term cannot vanish under non-uniformity: if all $\sigma_j$ are equal, $\tilde W\tilde W^T=\frac{n}{m}\boldsymbol I$, and all $h_i=\frac{m}{n}$. Non-uniform $h_i$ thus requires non-uniform $\sigma_j$, and $\mathrm{Var}^V_i\sigma_j^2$ may enter the loss. Its effect is beyond the scope of this paper. Second, the mean field approximation in Eq.~(\ref{eq:append-nonuni-xtalk-noise-variance-meanfield}) assumes that the overlaps of feature $i$ are spread over all other features regardless of their firing probabilities, so that the crosstalk noise depends on other features only through $\langle s^2p\rangle$. In trained models, the organization of representation vectors is correlated with firing probabilities. For example, \citet{liu_superposition_2025} found that the representation vectors of more frequent features tend to be ETF-like, with nearly equal and small overlaps among themselves. The crosstalk noise on a feature then depends on which features it overlaps with, rather than only on the mean.

Finally, the perturbation analysis requires $|\delta p_i|\ll p_0$ for all features. The power-law firing probabilities used in Sec. \ref{sec:non-unif-indep} do not satisfy this condition in general. For $p_i\propto i^{-\epsilon_p}$, $\delta p_i/p_0\approx\epsilon_p(\log\frac{n}{i}-1)$, so the condition requires $\epsilon_p\log n\ll 1$. For $n=10000$ and $\epsilon_p=0.2$ as in Fig. \ref{fig:non-uniformity-loss}, $\epsilon_p\log n\approx 1.8$, and the firing probability of the most frequent feature is about five times $p_0$. Therefore, we do not expect Eq.  (\ref{eq:append-susceptibility-sbh}) and (\ref{eq:append-correction-loss-nonuni}) to give correct values in this setup. They only explain the trends: the representation adapts to non-uniformity, and the loss decreases with $\epsilon_p$.

\section{Toy model training}
\label{sec:toy-model-training}

In this section, we explain how we train the toy model to obtain raw data. We carry out toy model experiments using PyTorch. Hyperparameters we use are given as follow:
\begin{itemize}
\item Data dimension $n$ (total number of features): varies from 1000 to 65536. In the main text, we only show results at $n=10000$.
\item Model width $m$: varied from 10 to 16384. In each case, the model width $m$ is always less than $n$.
\item Batch size: set to be $n$ in each case.
\item Total training steps: 30000 (tested up to 50000, which does not change any result; see Sec. \ref{sec:train-conv-exper})
\item Optimizer: Adam~\citep{kingma_adam_2017} with the default PyTorch settings ($\beta_1 = 0.9$, $\beta_2 = 0.999$, $\epsilon = 10^{-8}$, no weight decay).
\item Learning rate: dynamically adjusted using cosine decay scheduling
  \begin{equation}
    \label{eq:learning-rate-expression}
    l(t)=\left \{
  \begin{array}{ll}
    l_{{\rm max } } {t \over t_{{\rm 1 } }}, & t<t_{{\rm 1 } }, \\
    l_{{\rm fnl } } + (l_{{\rm max } } - l_{{\rm fnl } }) \cdot {1\over 2} \left ( 1 + \cos \pi {t - t_{{\rm 1 } } \over t_{{\rm 2 } } - t_{{\rm 1 } } } \right ), & t_1\le t<t_{{\rm 2 } }, \\
    l_{{\rm fnl } }, & t\ge t_{{\rm 2 } },
  \end{array}
\right .
  \end{equation}
  where warm-up time $t_1=2000$, step number of learning rate decay $t_2=20000$, and final learning rate $l_{{\rm fnl } }=0.0002$, the maximal learning rate $l_{{\rm max } }=20l_{{\rm fnl } }$.
\item Initialization of $\boldsymbol{W}$: each representation vector $\boldsymbol{w}_i$ is drawn independently and uniformly from the unit sphere in $\mathbb{R}^m$. The initial configuration is thus a random projection frame with $s_i = 1$ for all features.
\item Initialization of $\boldsymbol{b}$: $\boldsymbol{b} = \boldsymbol{0}$.
\end{itemize}
A new batch is randomly sampled from the data distribution at every step, so the model is trained on the input ensemble instead of a fixed dataset. Training is performed using Nvidia L40S, with floating-point precision (FP32).

\section{Details of experimental results on the toy model}
\label{sec:deta-exper-results}

In this section, we show more experimental results on the toy model of superposition. All models are trained as described in Appendix \ref{sec:toy-model-training}. The loss $\mathcal L$ is averaged over the last 5000 training steps. Representation parameters $\boldsymbol w_i$ and $b_i$ are measured after training; parameters $s_i$, $h_i$ and $r$ are calculated from $\boldsymbol w_i$.

\subsection{Uniform independent features}
\label{sec:det-exp-results-unif-indep-feat}

\subsubsection{Training convergence and experimental errors}
\label{sec:train-conv-exper}

\begin{figure}[htb]
\begin{center}
\includegraphics[width=\textwidth]{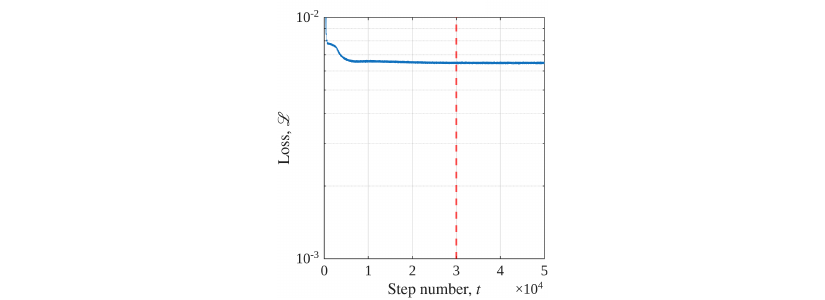}
\end{center}
\caption{Time sequences of trained loss. The loss converges before $t=3\times 10^4$. 
  $\bar E=64,n=10000,\sigma _v={1\over \sqrt{ 3 }}, m=200$.}
\label{fig:append-Loss_curves}
\end{figure}

Fig. \ref{fig:append-Loss_curves} shows an example loss sequence during training. The loss $\mathcal L$ reaches a plateau before the learning rate reaches its final value at $t_2 = 20000$, and remains constant within the last 5000 steps, over which we average the loss. Training longer, up to $5\times 10^4$ steps, does not change the results.

\begin{figure}[htb]
\begin{center}
\includegraphics[width=\textwidth]{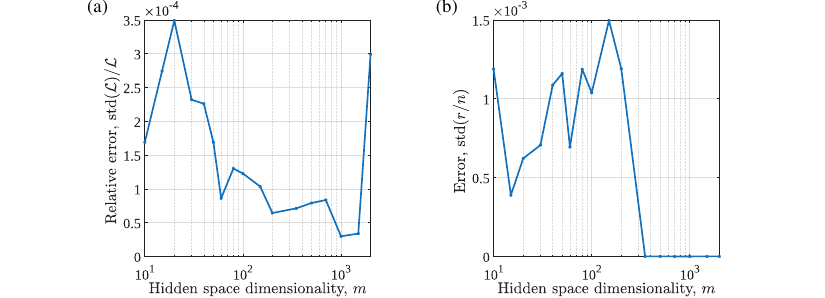}
\end{center}
\caption{(a) Relative error of the loss $\mathcal L$. (b) Error of $r/n$. The typical value of $r/n$ is of order 1. Since we focus on the transition, where $r/n$ reaches 1, we report the absolute rather than the relative error of $r/n$. For both panels 10 seeds are used. $\bar E=50,n=10000,\sigma _v={1\over \sqrt{ 3 }}$.}
\label{fig:append-Errors}
\end{figure}

We repeat several parameter settings with 10 random seeds. In all tested cases, the errors (seed-to-seed standard deviations) of $\mathcal L$ and $r/n$ are too small to be visible in the figures (Fig. \ref{fig:append-Errors}). We therefore use a single seed for most settings and omit error bars in most figures.

Near a continuous phase transition, statistical mechanics suggests that fluctuations of the order parameter $r/n$ (the susceptibility) and of the loss $\mathcal{L}$ (which plays the role of an effective energy) grow. In the thermodynamic limit, these may diverge at the critical point. At our resolution, this effect is weak: the standard deviation of $r/n$ increases only mildly below the transition, and the relative standard deviation of $\mathcal{L}$ shows no peak near it (Fig.~\ref{fig:append-Errors}). It does not affect our results, since we do not locate the critical point precisely; we only extract the trend of $m^*$ from the phase diagrams.

\subsubsection{Effect of $\bar v$ and $\sigma _v$}
\label{sec:effect-sigma-_v}

In this section, we discuss the effect of $\bar v$ and $\sigma _v$. The mean activation $\bar v$ trivially acts as a scaling factor of the model. Under the rescaling
\begin{align}
  \bar v \to & C\bar v, \\
  \sigma _v^2 \to & C^2 \sigma _v^2,\\
  \boldsymbol W \to & \boldsymbol W, \\
  \boldsymbol b\to & C \boldsymbol b,
\end{align}
with positive $C$, the loss $\mathcal L \to C^2 \mathcal L$. So in this paper, without loss of generality, we fix $\bar v=1$.

\begin{figure}[htb]
\begin{center}
\includegraphics[width=0.9\textwidth]{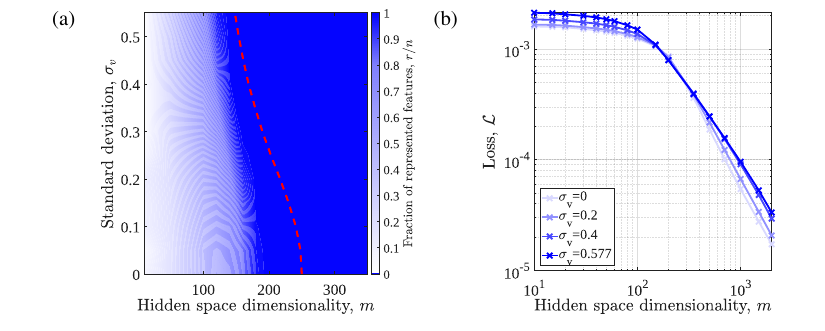}
\end{center}
\caption{(a) Phase diagram of $r/n$ as a function of $m$ and $\sigma _v$. The red dashed curve is the critical width $m^{*}$ predicted by partial RP approximation. The critical width $m^{*}$ decreases with $\sigma _v$. (b) The $\mathcal L$-$m$ at different $\sigma _v$. The loss $\mathcal L$ increases with $\sigma_v$. $\bar E=16,n=10000$.}
\label{fig:append-sigma-v-effect}
\end{figure}

With $\bar v = 1$, the activation strength $v$ is drawn from $U[1-\sqrt{3}\sigma_v,\ 1+\sqrt{3}\sigma_v]$. Nonnegative activations require $\sigma_v \le 1/\sqrt{3}$, so we vary $\sigma_v$ from 0 to $1/\sqrt{3}$, where $v \sim U[0, 2]$ as in the main text. The activation variance $\sigma _v^2$ changes the loss $\mathcal L$ and the critical width $m^{*}$. Experiments show that $\mathcal L$ increases with $\sigma _v^2$ (Fig. \ref{fig:append-sigma-v-effect}b). This trend is captured by the partial RP approximation, see Eq. (\ref{eq:append-loss-not-represented}) and (\ref{eq:append-LpRP-no-sigmaE-small-noise}). The critical width $m^{*}$ decreases with $\sigma _v$ (Fig. \ref{fig:append-sigma-v-effect}a). The partial RP approximation numerically captures this trend, but it cannot be described by asymptotic expressions derived in Sec. \ref{sec:asympt-behav-part}. This is mainly a consequence of the nonlinearity of the loss.

\subsubsection{Two constrained representations: equal-norm frame and random projection frame}
\label{sec:two-constr-repr}

\begin{figure}[htb]
\begin{center}
\includegraphics[width=\textwidth]{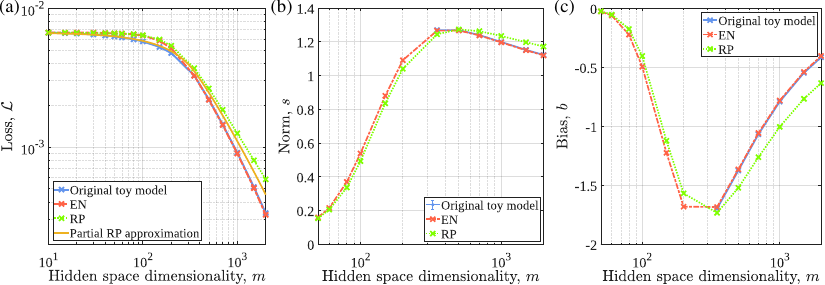}
\end{center}
\caption{Losses and representation parameters of different models. (a) Losses of different models. All loss curves have the same trend and similar bending. (b) The vector norms of different models. For the original toy model, the mean norm of all features is shown, and only the data in full-representation phase ($350\lesssim m$) is shown. The error bar is too small to be seen. (c) The biases of different models. For the original toy model, the mean bias of all features is shown; only the data in full-representation phase ($350\lesssim m$) is shown. $\bar E=50, n=10000, \sigma _v={1\over \sqrt{ 3 }}$.}
\label{fig:append-loss-scale-bias-diff-setup}
\end{figure}

In experiments and in partial RP approximation, features are divided into two groups: represented and unrepresented groups. To isolate the behavior of the represented group, we introduce two variant models with constrained representations based on the original toy model in Sec. \ref{sec:toy-model-its}:
\begin{enumerate}
\item The equal-norm (EN) model: During training, all vectors are forced to have equal norms. So the representation is forced to be an equal-norm frame.

  We parametrize each representation vector as $\boldsymbol{w}_i = s\, \hat{\boldsymbol{w}}_i / \|\hat{\boldsymbol{w}}_i\|$, where $\hat{\boldsymbol{w}}_i \in \mathbb{R}^m$ are unconstrained trainable vectors and $s$ is a single trainable scalar shared by all features. The directions of the representation vectors are thus trained freely, while all norms equal $s$ at every step, so the representation is always an equal-norm frame. The biases $b_i$ are trained independently for each feature. We initialize $\hat{\boldsymbol{w}}_i$ from a standard Gaussian, $s = 1$ and $\boldsymbol{b} = \boldsymbol{0}$, so the initial configuration is the same random projection frame as in the original model (Sec. \ref{sec:toy-model-training}).
  
\item The random projection (RP) model: During training, directions of all representation vectors are not updated and norms of them are forced equal. Furthermore, the bias of all features are forced equal. So there are only two trained parameters: norm $s$ and bias $b$. The vector configuration remains the initial random projection frame (see Sec. \ref{sec:toy-model-training}).
\end{enumerate}
The EN model is the realization of ideal full representation, and the RP model is a further simplified case. Neither of them captures the phase transition from partial to full representation. Experiments show that, in full-representation phase, the loss $\mathcal L$ and representation parameters $b$ and $s$ of the EN model are very close to those of the original model (Fig. \ref{fig:append-loss-scale-bias-diff-setup}). This result is also consistent with the observation that in full-representation phase, all $s_i$ (and $b_i$) are highly clustered. Therefore, in this section and the following section \ref{sec:finite-size-effect}, we treat the EN model and the original toy model in full-representation phase as equivalent.

\begin{figure}[htb]
\begin{center}
\includegraphics[width=\textwidth]{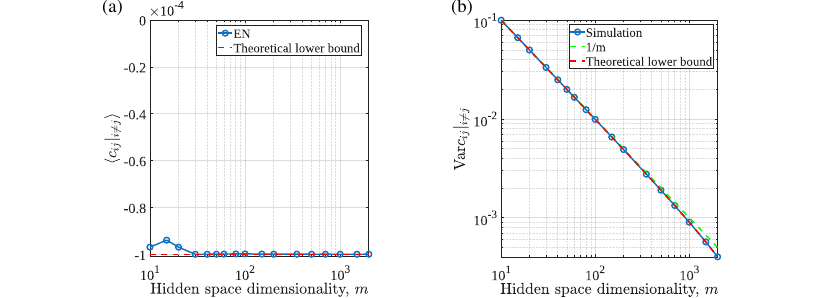}
\end{center}
\caption{(a) The mean value of  $c_{ij}|_{i\ne j}$ in the EN model. It reaches the theoretical lower bound $-{1\over n-1}$. (b) The variance of  $c_{ij}|_{i\ne j}$ in the EN model. It reaches the theoretical lower bound ${n-m \over m(n-1)}$. The curve $1/m$ is plotted out as a comparison. $\bar E=10,n=10000,\sigma _v={1\over \sqrt{ 3 }}$.}
\label{fig:append-mean_var_cij}
\end{figure}

We measure the mean value and the variance of $c_{ij}|_{i\ne j}$ in the trained EN model. For all parameters we tested, the mean value and the variance of $c_{ij}|_{i\ne j}$ both reach the theoretical lower bound (Fig. \ref{fig:append-mean_var_cij}) in the EN model, hence the trained representation is nearly a tight frame.

\begin{figure}[htb]
\begin{center}
\includegraphics[width=\textwidth]{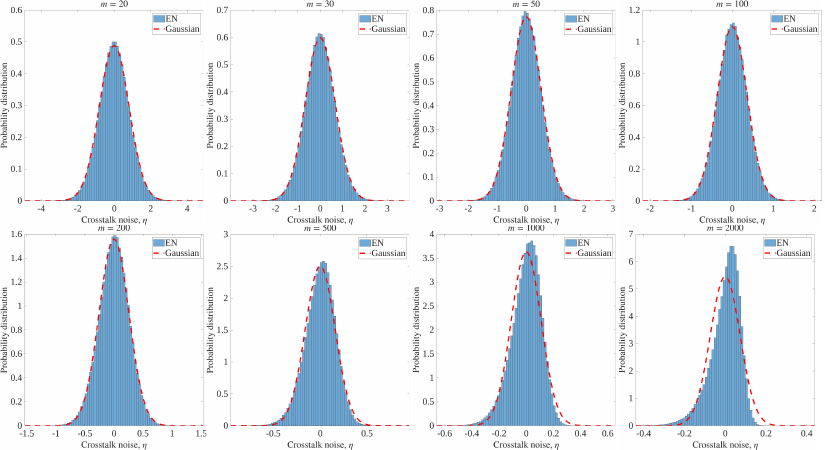}
\end{center}
\caption{The distribution of the crosstalk noise $\eta $ reconstructed from $c_{ij}$ of the EN model. The red dashed line is the pdf of Gaussian distribution (\ref{eq:xtalk-noise-approx-distribution}). $\bar E=10,n=10000,\sigma _v={1\over \sqrt{ 3 }}$.}
\label{fig:append-EN_xnoise_distribution}
\end{figure}

\begin{figure}[htb]
\begin{center}
\includegraphics[width=\textwidth]{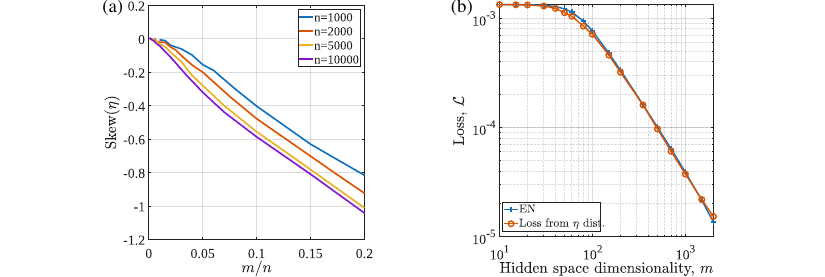}
\end{center}
\caption{(a) The skewness of $\eta $ at different $n$. The skewness decrease with $m/n$ from 0. $\bar E=10, \sigma _v={1\over \sqrt{ 3 }}$. (b) The trained loss of the EN model and the mean field loss calculated by applying the reconstructed $\eta $ distribution to Eq. (\ref{eq:append-loss-after-mean-field}). They are highly consistent. $\bar E=10,n=10000,\sigma _v={1\over \sqrt{ 3  }}$.}
\label{fig:append-Skewness_reconst_loss}
\end{figure}

Using the $c_{ij}$ measured in the experiments, we can also reconstruct the distribution of crosstalk noise $\eta $ in the EN model: To generate a crosstalk noise value, we sample $\bar E$ elements from the pool of $c_{ij}|_{i\ne j}$ and multiply them each by a random $v_i$ drawn from the distribution of activation, and then add them up. (Here we fix $E=\bar E$.) The reconstructed crosstalk noise distribution is close to the Gaussian approximation (\ref{eq:xtalk-noise-approx-distribution}) at small $m$, but deviates from Gaussian distribution as $m$ increases (Fig. \ref{fig:append-EN_xnoise_distribution}).
At large $m$, the distribution of $\eta$ becomes asymmetric (Fig.~\ref{fig:append-EN_xnoise_distribution}). Since its variance is fixed at the Welch bound (Fig. \ref{fig:append-mean_var_cij}b), the trained frame can lower the loss only by changing the shape of the distribution. On inactive represented features (which is much more than active represented features), the output is $[s^2\eta + b]_+$ with $b < 0$, so only the positive tail $\eta > -b/s^2$ leaks through the ReLU and bias and contributes to the loss. Correspondingly, the trained distribution is negatively skewed, with a lighter positive tail than the Gaussian distribution of the same variance. This suppresses the leakage, so the Gaussian approximation (\ref{eq:xtalk-noise-approx-distribution}) overestimates the loss. This is the main origin of the slightly higher predicted loss in the full-representation phase (Fig.~\ref{fig:analytic-solution}d). We use the skewness to quantify the deviation from Gaussianity. Empirically, it depends approximately on $m/n$ and vanishes as $m/n \to 0$ (Fig.~\ref{fig:append-Skewness_reconst_loss}a). We therefore regard the loss gap as a finite size effect, which disappears in the limit $1 \ll \bar E \ll n$ (Sec.~\ref{sec:finite-size-effect}). Plugging the reconstructed $\eta $ distribution into Eq. (\ref{eq:append-loss-after-mean-field}), we obtain a mean-field loss that matches the trained loss of the EN model (Fig. \ref{fig:append-Skewness_reconst_loss}b). This indicates that the mean field approximation is not the main cause of approximation error; the error mainly arises in using Gaussian distribution to approximate the crosstalk noise. We expect all these results to hold for the original toy model in full-representation phase too.

\begin{figure}[htb]
\begin{center}
\includegraphics[width=0.3\textwidth]{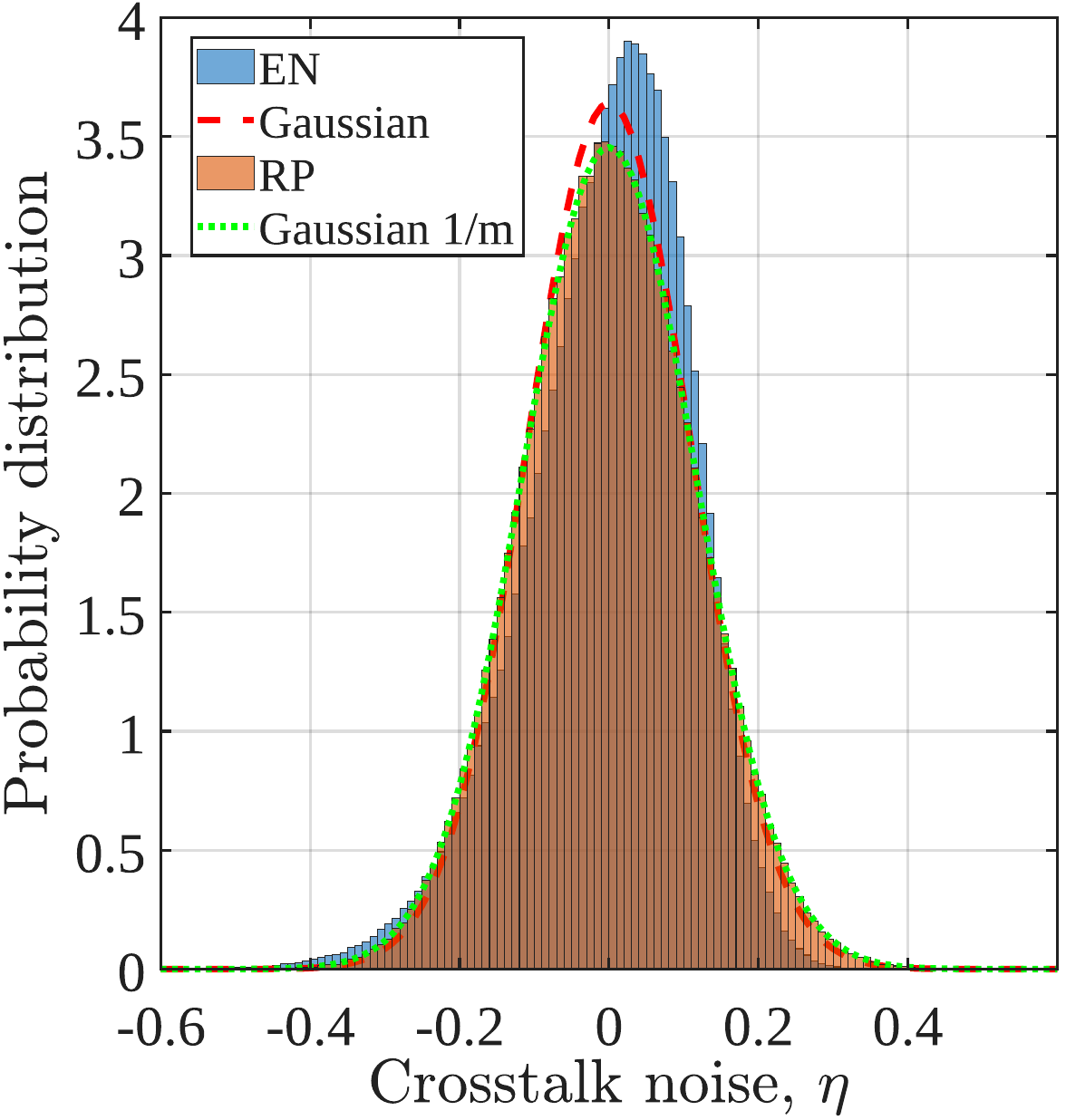}
\end{center}
\caption{The crosstalk noise distribution of the RP model and the EN model. The red dashed curve is the pdf of Gaussian distribution (\ref{eq:xtalk-noise-approx-distribution}), and the green dashed curve is the pdf of Gaussian distribution with variance $\bar E/m\cdot (\bar v^2+\sigma _v^2)$. The crosstalk noise distribution of the RP model is highly consistent with the Gaussian distribution with variance $\bar E/m\cdot (\bar v^2+\sigma _v^2)$.
  $\bar E=10,m=1000,n=10000,\sigma _v={1\over \sqrt{ 3 }}$.}
\label{fig:append-Xnoise_RP_vs_EN}
\end{figure}

The crosstalk noise distribution of the RP model is highly Gaussian, with a variance ${\bar E\over m}(\bar v^2+\sigma _v^2)$ rather than $\bar E{n-m\over m(n-1)}(\bar v^2+\sigma _v^2)$, as shown in Fig. \ref{fig:append-Xnoise_RP_vs_EN}. In the limit $m\ll n$, the two variances converge, and the crosstalk noise distribution of the RP model can be described by Eq. (\ref{eq:xtalk-noise-approx-distribution}). This is why we call our approximation the (partial) random projection approximation.

\subsubsection{Approaching the limit $1\ll \bar E \ll n$}
\label{sec:finite-size-effect}

Our asymptotic analysis of $\mathcal L_{{\rm rep } }$ lies in the limit $1\ll \bar E\ll n$. To approach this limit in experiments, we first  increase $n$ at fixed $\bar E\over n$ , then decrease $\bar E\over n$. To isolate the represented features, we use the EN model (Sec. \ref{sec:two-constr-repr}) instead of the original toy model, and compare its result with predicted $\mathcal L_{{\rm rep } }$ given by Eq. (\ref{eq:append-ind-binom-RP-loss}).

\begin{figure}[htb]
\begin{center}
\includegraphics[width=\textwidth]{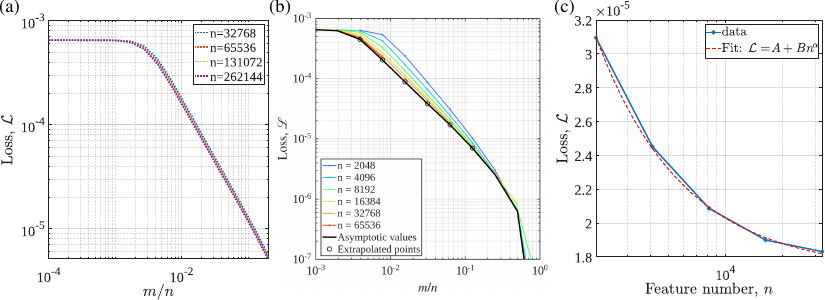}
\end{center}
\caption{(a) The represented loss $\mathcal L_{{\rm rep } }$ predicted by Eq. (\ref{eq:append-ind-binom-RP-loss}) at fixed $\bar E/n$. Loss curves collapse after rescaling $m$ to $m/n$. (b) The EN model loss $\mathcal L_{{\rm EN } }$ at different $n$ and its asymptotic value in $n\to \infty $. For each $m/n$, we fit $\mathcal L_{{\rm EN } }(n)$ by $\mathcal L_{{\rm EN } }=A+Bn^{\alpha }$. If the fit matches the loss, we use $A$ from fitting as the asymptotic value, otherwise we use the minimal measured loss as the experimental asymptotic value (an upper bound on the real asymptotic value). Circles mark the points based on fit result. (c) $\mathcal L_{{\rm EN } }$ and the fit result at $m/n=1/16$. $\bar E/n=1/2048$, $\sigma _v={1\over \sqrt{ 3 }}$.}
\label{fig:append-Loss_collapse_large_n}
\end{figure}

At fixed $\bar E/n$, loss curves of different $n$ collapse when $n$ is large enough. For the predicted $\mathcal L_{{\rm rep } }$, we scale $n$ up and confirm the collapse of $\mathcal L_{{\rm rep } }$-$m/n$ curves (Fig. \ref{fig:append-Loss_collapse_large_n}a). For the EN model, the maximal $n$ we can test is limited by our computational resources, hence we cannot see loss curves collapsing directly. However, by fitting $\mathcal L_{{\rm EN } }(n)$ at fixed $m/n$, we find that in a large range of $m/n$, $\mathcal L_{{\rm EN } }$ converges to an asymptotic value in the limit $n\to \infty $ (Fig. \ref{fig:append-Loss_collapse_large_n}bc).

\begin{figure}[htb]
\begin{center}
\includegraphics[width=\textwidth]{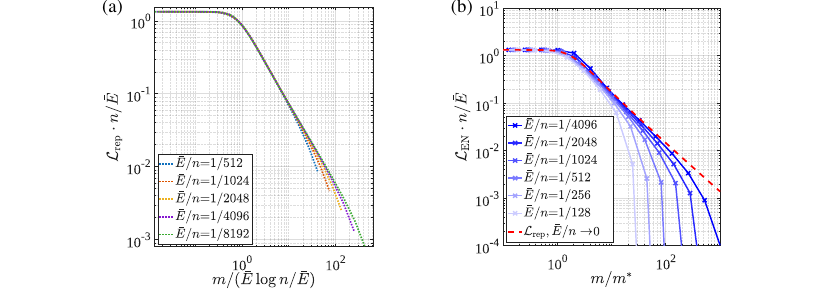}
\end{center}
\caption{(a) The represented loss $\mathcal L_{{\rm rep } }$ predicted by Eq. (\ref{eq:append-ind-binom-RP-loss}) at different $\bar E/n$. $n$ is large enough so that the large-$n$ collapse happens. (b) The asymptotic values of the EN model loss $\mathcal L_{{\rm EN } }$ at different $\bar E/n$. Experimental asymptotic values are obtained by the extrapolation described in Fig. \ref{fig:append-Loss_collapse_large_n}. The red dashed curve is the limiting value $\lim_{\bar E/n\to 0} \lim_{n\to \infty }\mathcal L_{{\rm rep } }$ computed numerically from Eq. (\ref{eq:append-ind-binom-RP-loss}). $\sigma _v={1\over \sqrt{ 3 }}$.}
\label{fig:append-Loss_collapse_small_Ern}
\end{figure}

We then decrease $\bar E/n$ to approach the limit $1\ll \bar E\ll n$. Based on the result of the original toy model, we rescale the loss $\mathcal L$ to $\mathcal L\cdot n/\bar E$, and the width $m$ to  $m/(\bar E \log {n\over \bar E})$ or equivalently $m/m^{*}$. For $\mathcal L_{{\rm rep } }$ predicted by Eq. (\ref{eq:append-ind-binom-RP-loss}), the loss curves collapse well at small $m$; at large $m$, due to the finite size effect (as $m$ reaches $n$, the loss decays toward 0), the loss curves deviate (Fig. \ref{fig:append-Loss_collapse_small_Ern}a). The asymptotic values of $\mathcal L_{{\rm EN}}$ show similar behaviors (Fig. \ref{fig:append-Loss_collapse_small_Ern}b). As $\bar E/n$ approaches 0, the asymptotic values of $\mathcal L_{{\rm EN}}$ converge to the predicted limiting value $\lim _{\bar E/n\to 0}\lim_{n\to \infty }\mathcal L_{{\rm rep } }$. We therefore conclude that in the limit $1\ll \bar E \ll n$, the EN model loss $\mathcal L_{{\rm EN } }$ converges to the predicted $\mathcal L_{{\rm rep } }$.

All analyses in this section use the EN model. Nevertheless, all results hold for the original toy model in the full-representation phase (see Sec. \ref{sec:two-constr-repr}). We also expect these results to be valid for $\mathcal L_{{\rm rep } }$ in partial-representation phase, because $s_i$ and $b_i$ of represented features are highly clustered in that phase. Together with Sec. \ref{sec:two-constr-repr}, this shows that the gap between the partial RP prediction and the trained loss in the full-representation phase (Fig. \ref{fig:analytic-solution}) is a finite size effect: it arises from the non-Gaussian crosstalk noise at finite $m/n$, and vanishes in the limit $1 \ll \bar E \ll n$.

\subsection{Exchangeable correlated features}
\label{sec:det-exp-results-exch-corr-feat}

\subsubsection{Generation of exchangeable correlated features}
\label{sec:gener-exch-corr}

In this section, we describe the method to generate $\boldsymbol u$ for exchangeable correlated features. The generation consists of:
\begin{enumerate}
\item Generating $E$, the number of active features in the input, following a discrete distribution $\{P_E\}$, where $P_E$ is the probability that $E$ active features occur.
\item Randomly selecting $E$ features uniformly at random. Selected features are set active, and the rest are inactive.
\end{enumerate}

The probability array $\{P_E\}$ is tunable, and is not uniquely determined by its mean $\bar E$ and variance $\sigma _E^2$. We restrict ourselves to the setup of Pólya–Eggenberger urn \citep{johnson_urn_1977}. Under this setup, $\sigma _E$ has four regimes with different $\{P_E\}$ implementation. At $\sigma _E=0$, the distribution degenerates to a delta at $\bar E$:
\begin{equation}
P_E=\delta _{E,\bar E}.
\end{equation}
At $\sigma _E = \sigma _E^{\rm bin }=\sqrt{ \bar   E(1-{\bar E\over n})}$, the distribution is just binomial distribution:
\begin{equation}
P_E = \binom{n}{E}\cdot \left (\bar E\over n\right )^E\left (1-{\bar E\over n}\right )^{n-E}.
\end{equation}
Features are activated independently in this case. In the range $0<\sigma _E<\sigma _E^{{\rm bin } }$, the activation number $E$ follows hypergeometric distribution:
\begin{equation}
	P_E={\binom{K}{E} \binom{N-K}{n-E} \over \binom{N}{n}},~~~{\rm for~ } E\in \{ \max(0,n+K-N),...,\min(n,K) \},
\end{equation}
where
\begin{align}
	N&\equiv {n-(\sigma _E/\sigma _E^{\rm bin } )^2 \over 1-(\sigma _E/\sigma _E^{\rm bin } )^2},\\
	K&\equiv {\bar E\over n}N.
\end{align}
In the range $0<\sigma _E^{{\rm bin } }<\sigma _E$, the activation number $E$ follows beta-binomial distribution:
\begin{equation}
P_E=\binom{n}{E} {B (E+\alpha ,n-E+\beta ) \over B(\alpha ,\beta )},
\end{equation}
where
\begin{align}
  B(x,y)&={\Gamma (x) \Gamma (y) \over \Gamma (x+y)},\\
  \alpha  &={\bar E \over n}{{(\sigma _E/\sigma _E^{\rm bin } )^2-n \over 1-(\sigma _E/\sigma _E^{\rm bin } )^2}},\\
	\beta   &=\left (1-{\bar E \over n}\right ){{(\sigma _E/\sigma _E^{\rm bin } )^2-n \over 1-(\sigma _E/\sigma _E^{\rm bin } )^2}}.
\end{align}

Following this setup, the possible $\sigma _E$ ranges from $0$ to $\sqrt{ \bar E (n-\bar E) }$. We restrict $\sigma _E$ to be of order $\bar E$.

\subsubsection{Effect of $\sigma  _{E}$}

\begin{figure}[htb]
\begin{center}
\includegraphics[width=\textwidth]{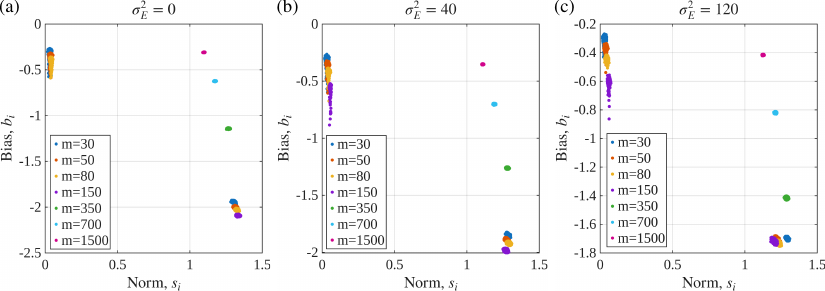}
\end{center}
\caption{The distribution of $s_i$ and $b_i$ of each feature. Sample points at small $m$ form two clusters while at large $m$ they form one cluster. Different panels are distributions at different $\sigma _E^2$. $\bar E=20,n=10000,\sigma _v={1\over \sqrt{ 3 }}$.}
\label{fig:append-norm_bias_clusters_Exc_corr}
\end{figure}

\begin{figure}[htb]
\begin{center}
\includegraphics[width=\textwidth]{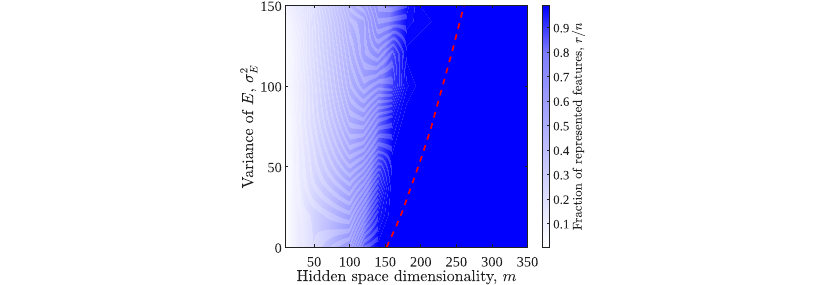}
\end{center}
\caption{Phase diagram of $r/n$ as a function of $m$ and $\sigma _E^2$. The red dashed curve is the critical width $m^{*}$ predicted by partial RP approximation. The critical width $m^{*}$ increases with $\sigma _E^2$. $\bar E=20,n=10000,\sigma _v={1\over \sqrt{ 3 }}$.}
\label{fig:append-Exc_corr-crit-m}
\end{figure}

\begin{figure}[htb]
\begin{center}
\includegraphics[width=\textwidth]{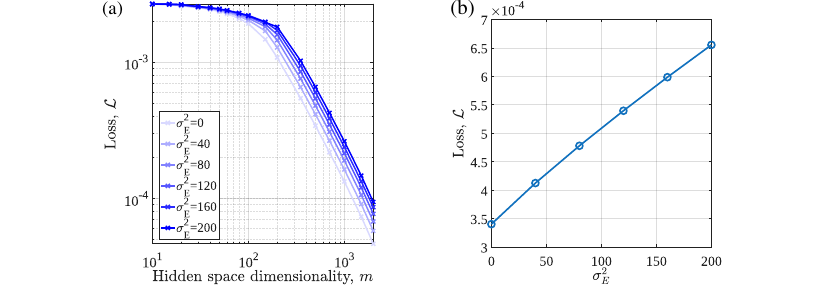}
\end{center}
\caption{(a) The loss $\mathcal L$ as a function of $m$ at different $\sigma _E$. In full-representation phase, the loss $\mathcal L$ increases with $\sigma_E$. (b) The loss $\mathcal L$ at $m=500$, which is greater than the critical width. In full-representation phase, the loss $\mathcal L$ is approximately linear in $\sigma _E^2$. $\bar E=20,n=10000,\sigma _v={1\over \sqrt{ 3 }}$.}
\label{fig:append-Exc_corr-loss}
\end{figure}

We test various $\bar E$ and $\sigma _E$, with $\sigma _E^2\le 200$, and show $\bar E = 20$ as a representative example; other values of $\bar E$ give qualitatively the same results. In the parameter range we test, the phase transition from partial representation to full representation remains valid (Fig. \ref{fig:append-norm_bias_clusters_Exc_corr}). Toy model experiment shows that the critical width $m^{*}$ grows with $\sigma _E$. This trend is captured by partial RP approximation (Fig. \ref{fig:append-Exc_corr-crit-m}). In the full-representation phase, the loss grows with $\sigma _E^2$ approximately linearly at small $\sigma _E^2$, as predicted in Sec. \ref{sec:deta-deriv-loss-exch-corr-feat}; at larger $\sigma_E^2$, the growth becomes slightly nonlinear (Fig. \ref{fig:append-Exc_corr-loss}). In the partial-representation phase, since the loss is dominated by unrepresented features, it is insensitive to $\sigma _E$.

We expect the representation behavior will be qualitatively different at large enough $\sigma _E$. For example, when $\sigma _E$ reaches its upper bound, all features can only fire together, hence one vector is enough to represent the firing states of all features. If $\sigma _v$ also vanishes, a model with width 1 can recover all inputs. But the representation behavior at large $\sigma _E$ lies beyond the scope of this paper.

\subsection{Independent features with unequal firing probabilities}
\label{sec:det-exp-results-indep-feat-with}

\subsubsection{Applicable range of $\epsilon _p$ and bias behavior}

\begin{figure}[htbp]
\begin{center}
\includegraphics[width=\textwidth]{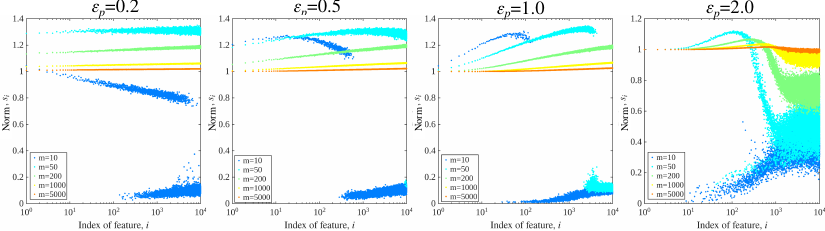}
\end{center}
\caption{Distributions of $s_i$ at different $\epsilon _p$. At small $\epsilon _p$, the distribution of $s_i$ show clear bimodality at small $m$; at large $\epsilon_p$, the two bands of $s_i$ merge at small $m$, and features cannot be divided into represented and unrepresented groups. $\bar E=1,n=10000,\sigma _v={1\over \sqrt{ 3 }}$.}
\label{fig:append-Applicable_ep_range}
\end{figure}

With power-law firing probabilities $p_i = A\, i^{-\epsilon_p}$, the prefactor is fixed by $\sum_i p_i = \bar E$, i.e., $A = \bar E / H_n(\epsilon_p)$ with $H_n(\epsilon_p) \equiv \sum_{i=1}^{n} i^{-\epsilon_p}$. Since $p_1 = A$ is a probability, we require $p_1 < 1$, i.e., $H_n(\epsilon_p) > \bar E$. Because $H_n(\epsilon_p)$ decreases with $\epsilon_p$, this condition sets an upper bound on $\epsilon_p$ at given $\bar E$ and $n$. For $n = 10000$, the bound is  $\epsilon_p < 0.995$ at $\bar E = 10$ and $\epsilon_p < 0.70$ at $\bar E = 50$, while at $\bar E = 1$ any $\epsilon_p$ is allowed. We therefore use $\bar E = 1$ to probe large $\epsilon_p$ (Fig. \ref{fig:append-Applicable_ep_range}).

The transition from partial to full representation persists only for moderate $\epsilon_p$ (Fig. \ref{fig:append-Applicable_ep_range}). At small $\epsilon_p$, the distribution of $s_i$ at small $m$ is clearly bimodal: frequent features form a band with $s_i$ of order 1, and rare features form a band with nearly vanishing $s_i$. As $\epsilon_p$ increases, the two bands approach each other. At large $\epsilon_p$ (e.g., $\epsilon_p = 2$), they merge even at small $m$: $s_i$ varies continuously with $i$, and features can no longer be divided into represented and unrepresented groups. In this regime, the partial representation and the partial RP approximation built on it no longer apply. Empirically, the transition remains well defined for $\epsilon_p \lesssim 1$.

\begin{figure}[htbp]
\begin{center}
\includegraphics[width=\textwidth]{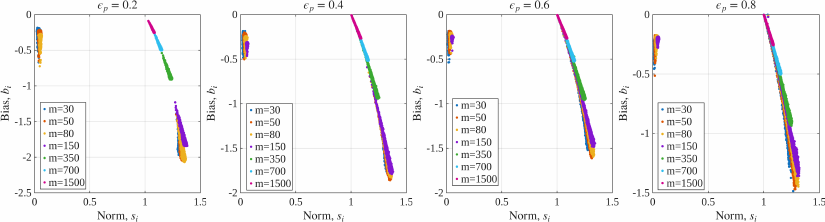}
\end{center}
\caption{The distribution of $s_i$ and $b_i$ of each feature. Sample points at small m form two clusters while at large m they form one cluster. Different panels are distributions at different $\epsilon _p$. $\bar E=10,n=10000,\sigma _v={1\over \sqrt{ 3 }}$.}
\label{fig:append-Norm_bias_clustering_nonuni}
\end{figure}

Within this range, the biases behave in the same way as the norms (Fig. \ref{fig:append-Norm_bias_clustering_nonuni}). At small $m$, the pairs $(s_i, b_i)$ form two clusters, corresponding to represented and unrepresented features; at large $m$, they merge into one. Unlike the uniform case (Fig. \ref{fig:transition}b), the represented cluster is elongated rather than point-like, reflecting the $p_i$-dependent norms and biases (Fig. \ref{fig:non-uniformity}). Along this cluster, $b_i$ decreases as $s_i$ increases, consistent with the perturbative result that more frequent features have smaller norms and larger biases (Eq. (\ref{eq:append-signs-of-susceptibilities})). The elongation grows with $\epsilon_p$.

\subsubsection{Comparison with the perturbation analysis}

We compare the perturbation analysis of Sec. \ref{sec:pert-analys-indep} with trained models. We only test the full-representation phase, where the baseline of Sec. \ref{sec:lowest-order-pert-to-part-RP} (all features represented) applies. The linear-response coefficients in Eq. (\ref{eq:append-susceptibility-sbh}) are evaluated at the uniform solution and depend only on $\bar E$, $m$, $n$ and $\sigma_v$, not on $\epsilon_p$; the data exponent only sets the deviations $\delta p_i$. It therefore suffices to test a single $\epsilon_p$. Here we show results at $\epsilon_p = 0.2,\sigma _v={1\over 3}$.

\begin{figure}[htbp]
\begin{center}
\includegraphics[width=0.3\textwidth]{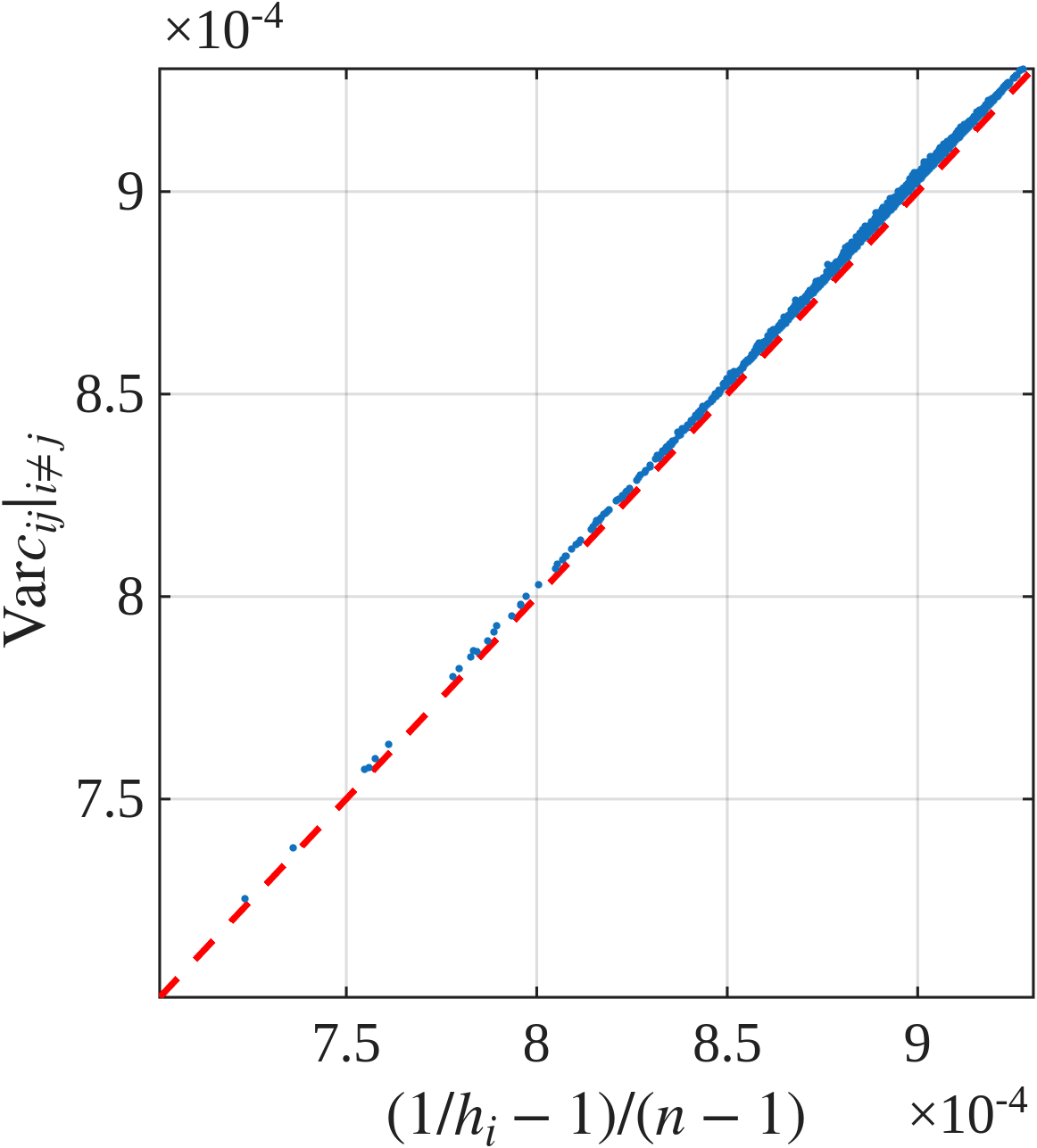}
\end{center}
\caption{Test of Eq. (\ref{eq:append-bridge-xtalk-leverage-score}) in the trained model. Each point is one feature. The horizontal axis is $({1\over h_i}-1)/(n-1)$, and the vertical axis is the variance of $c_{ij}$ over $i\ne j$. The red dashed line is $x=y$.
  $\bar E=10,n=10000,\sigma _v={1\over 3},\epsilon _p=0.2,m=1000$.}
\label{fig:Dist_1rHi_VarCij}
\end{figure}

We first test Eq.~(\ref{eq:append-bridge-xtalk-leverage-score}), which relates the overlaps of a feature to its leverage score. Keeping the constant term in Eq. (\ref{eq:append-accurate-crosstalk-overlap}) and dropping only the term $h_i \mathrm{Var}^V_i \sigma_j^2$, the sum of squared overlaps of feature $i$ is $1/h_i - 1$. Fig. \ref{fig:Dist_1rHi_VarCij} compares the measured variance of $c_{ij}$ over $j \neq i$
with $(1/h_i - 1)/(n-1)$ for each feature. The two agree across all features, so the neglected term is small and Eq.~(\ref{eq:append-bridge-xtalk-leverage-score}) holds in trained models under non-uniformity.

\begin{figure}[htbp]
\begin{center}
\includegraphics[width=\textwidth]{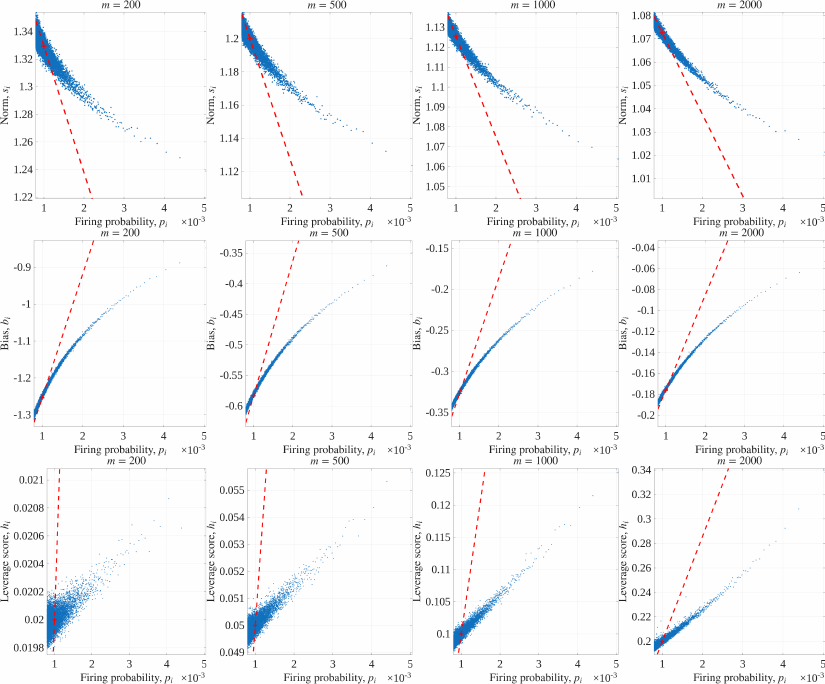}
\end{center}
\caption{Norms $s_i$ (top row), biases $b_i$ (middle row), and leverage scores $h_i$ (bottom row) as functions of the firing probability $p_i$ at different $m$. Each point is one feature. The red dashed lines are the linear-response predictions of Eq. (\ref{eq:append-susceptibility-sbh}) around the uniform solution. The predictions capture the signs of the responses but overestimate their magnitudes. $\bar E=10,n=10000,\sigma _v={1\over 3},\epsilon _p=0.2$.}
\label{fig:append-SBH_pert}
\end{figure}

We then compare the responses of $s_i$, $b_i$ and $h_i$ with Eq. (\ref{eq:append-susceptibility-sbh}) (Fig.~\ref{fig:append-SBH_pert}). At all tested widths, the signs of the responses agree with Eq. (\ref{eq:append-signs-of-susceptibilities}): $s_i$ decreases, while
$b_i$ and $h_i$ increase with $p_i$. The measured $h_i$ are also centered at the baseline value $h_0 = m/n$. The magnitudes, however, are overestimated. The measured responses are also nonlinear in $p_i$. This is expected, since $p_i$ reaches about $5p_0$, far outside the range $|\delta p_i| \ll p_0$ of the linear expansion (Sec. \ref{sec:disc-lowest-order}). The perturbation analysis thus captures the direction of the adaptation to non-uniformity, but not its magnitude.

\end{document}